\documentclass[preprint]{elsarticle}

\usepackage{graphicx}
\usepackage{amssymb}

\usepackage[caption=false]{subfig}
\usepackage{amsmath,stmaryrd}   
\usepackage{color}

\usepackage{hyperref}

\usepackage{caption}

\usepackage{dirtytalk}

\usepackage[ruled,vlined]{algorithm2e}

\newcommand{\aside}[1]{}

\usepackage{lineno}
\usepackage{multirow}

\journal{Sustainable Cities and Society}

\begin{document}

\begin{frontmatter}


\title{Urban feature analysis from aerial remote sensing imagery using self-supervised and semi-supervised computer vision}



\author[addressUoM]{Sachith Seneviratne}
\author[addressNL]{Jasper S. Wijnands}

\author[addressUoM]{Kerry Nice}
\author[addressUoM]{Haifeng Zhao}
\author[addressUoM]{Branislava Godic}
\author[addressPop,addressEPA]{Suzanne Mavoa}
\author[addressUoM,addressEng]{Rajith Vidanaarachchi}

\author[addressUoM,addressEng,addressPop]{Mark Stevenson}
\author[addressQUB]{Leandro Garcia}
\author[addressQUB]{Ruth F. Hunter}
\author[addressUoM]{Jason Thompson}

\address[addressUoM]{Transport, Health and Urban Design Research Lab, Melbourne School of Design, The University of Melbourne, Parkville VIC 3010, Australia}

\address[addressNL]{Royal Netherlands Meteorological Institute (KNMI), Utrechtseweg 297, De Bilt, The Netherlands}

\address[addressQUB]{Centre for Public Health, Queen’s University Belfast, Belfast, Northern Ireland}

\address[addressEng]{Faculty of Engineering and Information Technology, University of Melbourne, Melbourne, Australia}

\address[addressPop]{Melbourne School of Population and Global Health, University of Melbourne, Melbourne, Australia}

\address[addressEPA]{Environmental Public Health Branch, Environment Protection Authority Victoria, Melbourne, Australia. }

\begin{abstract}
Analysis of overhead imagery using computer vision is a problem that has received considerable attention in academic literature. Most techniques that operate in this space are both highly specialised and require expensive manual annotation of large datasets. These problems are addressed here through the development of a more generic framework, incorporating advances in representation learning which allows for more flexibility in analysing new categories of imagery with limited labeled data. First, a robust representation of an unlabeled aerial imagery dataset was created based on the momentum contrast mechanism. This was subsequently specialised for different tasks by building accurate classifiers with as few as 200 labeled images. The successful low-level detection of urban infrastructure evolution over a 10-year period from 60 million unlabeled images, exemplifies the substantial potential of our approach to advance quantitative urban research.
\end{abstract}


\end{frontmatter}

\section{Introduction}
\label{S:1}

Advances in deep learning methods \cite{schmidhuber2015deep} have enabled the analysis of very large datasets, including those containing  overhead and satellite imagery, in a fully automated manner. High-definition aerial imagery datasets are becoming increasingly available as a result of improved capture and storage techniques, as well as advances in processing power. Combined, this is enabling detailed analysis of higher resolution remote sensing scenes. The traditional deep learning process follows the steps of data collection, data labeling, model training and inference on unlabeled data to assign labels automatically to the unlabeled data.

Due to the sheer volume of available data, computer vision techniques are uniquely suited to efficiently process them for different tasks such as classification, object detection and semantic segmentation. In machine learning, supervised learning, which operates on labeled data to build a predictive model, has been extensively used to harness information from aerial imagery. Supervised learning techniques excel when provided with a large volume of labeled data. However, such data needs to be labeled manually which is labour-intensive and therefore expensive and difficult to scale. In contrast, unlabeled data such as satellite imagery is more freely available and exists in greater quantities. Several learning paradigms have investigated how to harness unlabeled data sources more efficiently including self-supervised learning and semi-supervised learning.

Recent advances in aerial imagery techniques have led to a rapid increase in the amount of overhead imagery available. This increase is led primarily by the higher resolution of imagery capture (for example - imagery captured at 10cm resolution would generate 100 times more data compared to imagery captured at 100cm (1m) resolution. However, in order to make use of this data, storage and processing power must also keep up. It is therefore imperative that analytical pipelines are capable of handling such data while maintaining key performance metrics such as analytical accuracy and speed.

High-resolution aerial imagery captures detailed urban characteristics,  enabling the potential identification of important urban features \cite{rs13040808} such as cycling infrastructure at scale. This work introduces methods for effectively exploring such large volumes of data (scaling up to 60 million images), using a much smaller labeled set of images (as few as 200 images). Methods leveraging self-supervision, semi-supervision are introduced, evaluated and deployed across 15 cities in Australia.

\subsection{Advances in neural network training techniques}

\subsubsection{Self-supervised representation learning}
\label{S2:selfsupervised}

Self-supervised learning extracts knowledge from unlabeled datasets by setting up a pretext task on which the model can be pretrained in a supervised manner~\cite{jing2020self}. In self-supervised workflows, the focus is on the intermediate representation that is learned by the self-supervision pretext task, rather than maximising prediction accuracy. This intermediate representation is used in downstream tasks such as object detection, with the expectation that the representation learned during the pretext task is robust from a semantic and structural perspective.

\begin{figure}[h]
    \subfloat[Solving jigsaw puzzles \cite{noroozi2016unsupervised}]{\setlength{\fboxsep}{0pt}\includegraphics[angle=0,scale=0.10]{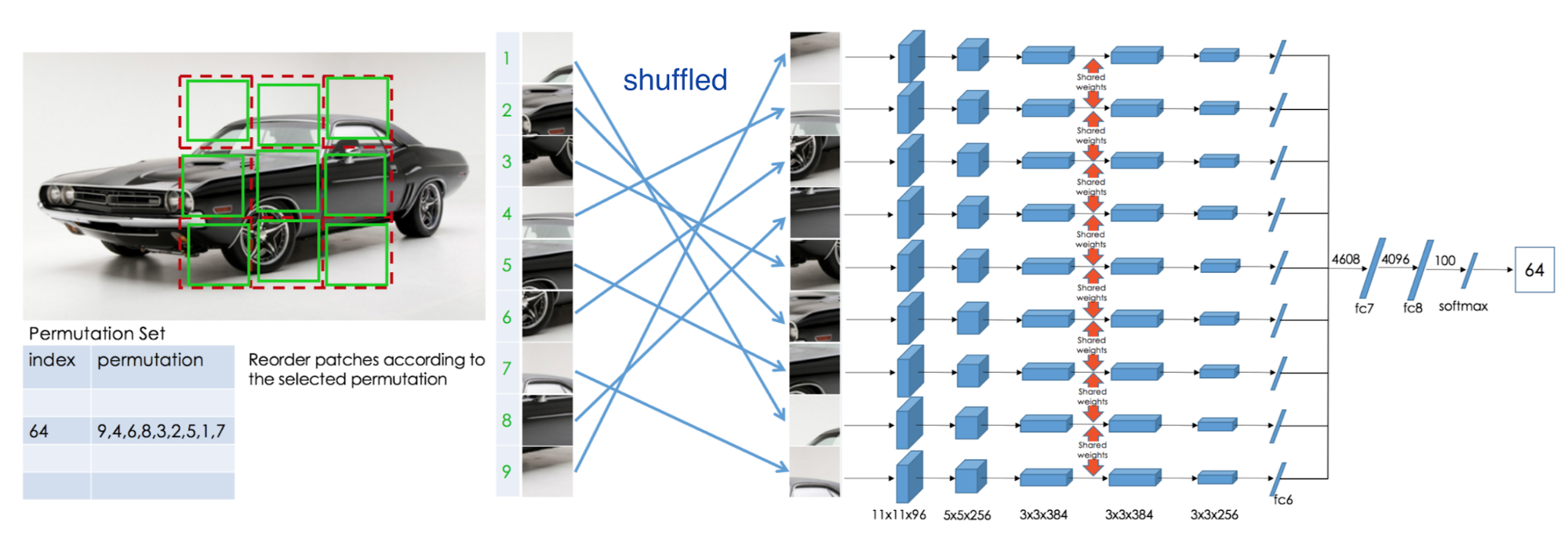}}\,
    \subfloat[Relative position \cite{doersch2015unsupervised}]{\setlength{\fboxsep}{0pt}\includegraphics[angle=0,scale=0.10]{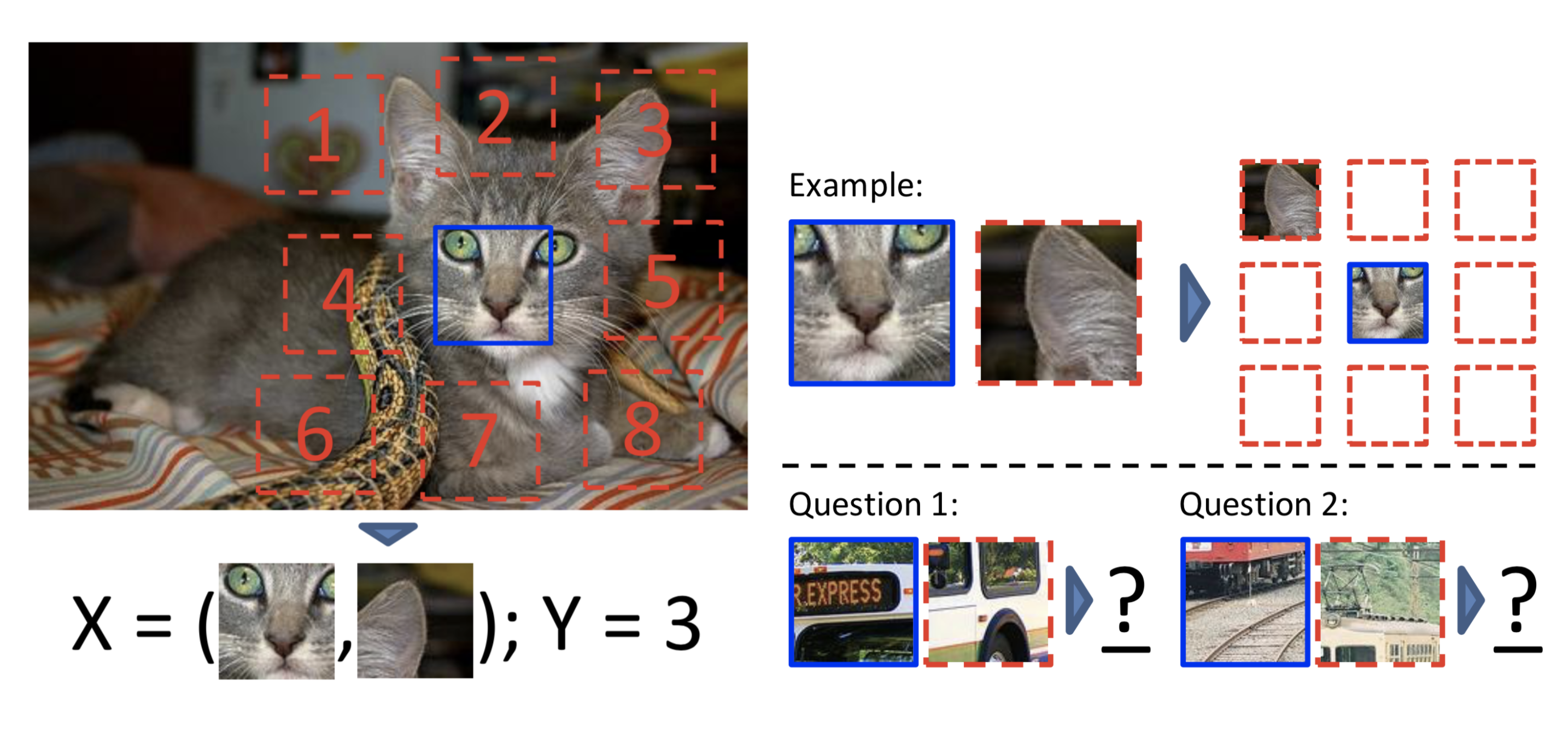}}
    \caption{Examples of pretext tasks. By performing such tasks, the neural network develops an initial understanding of the types of tasks it will be trained for in the future. This reduces the difficulty and magnitude of future training.}
    \label{fig:SelfSup}
\end{figure}

There is currently a large body of work focused on learning task-independent representations using these techniques. For example, \citet{noroozi2016unsupervised} formulated a jigsaw puzzle task by selecting several adjacent blocks of pixels. After shuffling the blocks, the model's task is to recover the correct spatial order (see Fig.~\ref{fig:SelfSup}a). This task requires high-level reasoning based on the objects and details visible in the image. Therefore, a model that excels in the pre-training task is likely to contain a useful representation of the image. Similarly, \citet{doersch2015unsupervised} designed the task of retrieving the relative position of a tile compared to a selected image section (see Fig.~\ref{fig:SelfSup}b).

Importantly, while self-supervised learning tends to reduce the labelling requirements for training neural networks, it does not provide a means for labelling large datasets. This is because self-supervised learning generally provides psuedo-labels for the model to build an initial representation of the world, which helps to reduce the number of labeled data points it needs to see to build a hypothesis about a particular category, but does not necessarily label data points related to those particular categories.

\subsubsection{Semi-supervised learning}

Semi-supervised learning corresponds to the class of machine learning techniques where a large amount of unlabeled data is available alongside a smaller collection of labeled data. These approaches attempt to use the small volume of labeled data to assign labels to the much larger volume of unlabeled data, in an iterative fashion. Thereby, the set of labeled data grows during analysis, leading to more accurate models.

Prior work has used semi-supervised approaches (also referred to as bootstrapping approaches in some research areas) to improve prediction accuracy of predictive models by generating more training data. However, very few operate in a fully automated manner. An early work in this paradigm of models learning by themselves is by \citet{yarowsky1995unsupervised}, who investigated the possibility of using labeled sentences coupled with unlabeled data to perform word-sense disambiguation. Several works also explored applicability of this technique in computer vision. For example, \citet{Cui_2016_CVPR} iteratively grew their dataset by merging in high-confidence predictions from their model. However, a manual vetting process was employed at each step. \citet{huang2015automatic} used morphology and colour-based indices using predefined formulae, as well as openly available information sources to generate training sets and classify images into the classes of buildings, roads, soil, water, shadow, and vegetation. A key issue with classification approaches is often that classes are assumed to be mutually exclusive. However, in aerial images of urban scenes, it is possible for roads, vegetation, soil, water and buildings to co-exist within the same image.

In general, the key improvements due to semi-supervised learning strategies can be incorporated at two levels:

\begin{itemize}
    \item \textbf{Model level} involves improvements incorporated into model training processes and focus on providing models with a more robust representation from fewer initially labeled image samples. 
    \item \textbf{Data level} involves improvements in the semi-supervised labelling process itself, allowing model-independent improvements via techniques such as heuristics and morphological feature extraction. 
\end{itemize}

Model level improvements generally include techniques which are useful even outside of the scope of semi-supervised learning as well. In fact, many of these techniques are used to improve supervised learning models. As examples, \citet{miyato2018virtual} use adversarial training,  \citet{siddharth2017learning} use disentangled feature learning and augmentation strategies such as RandAugment introduced in \citet{cubuk2020randaugment} are also commonly used.

Data level techniques operate outside of the scope of the model. These increase the probability of the model correctly labelling unlabeled image samples without human intervention. For example, \citet{kothari2020semisupervised} use unlabeled neighbourhood information to improve model performance.

As these two types of techniques apply at different levels, it is additionally possible to overlap them for potential combined improvements as well.

Most work in semi-supervised learning focuses on images captured in the horizontal perspective (images generated by cameras in non-aerial settings), due the abundance of labeled data which enables much easier model evaluation. By treating a large part of the dataset as unlabeled, it is still possible to easily evaluate model behaviour with small labeled datasets while also providing very robust accuracy, precision and recall metrics as required. Using an unlabeled dataset only enables the provision of estimates of such performance metrics, as the ground truth of a large part of the dataset is unknown. However, this type of analysis more accurately matches use of the technique with unlabeled datasets in the wild.

Table~\ref{Tab:Semisup_compare} contains a comparison of such techniques based on labeled set size, perspective and unlabeled set size. This comparison indicates reported results for the model using the lowest number of labeled images and not the number corresponding to the best results.

Many techniques compare performance based on a percentage of unlabeled data used as labeled data (for example, 1\% of data used as labeled data). However, this is not necessarily representative of annotation effort, which is a function of the absolute number of labeled images. As evaluation is primarily carried out using labeled data which is treated as unlabeled data (by hiding the label from the model), it is straightforward to do so. However, for use in the wild with a new unlabeled dataset, data annotation effort is often the limiting factor.  Additionally, most techniques report performance based on the training set size as a percentage of unlabeled/total set size, disregarding the labelling requirements for validation data. In this article, a major objective is to limit the total labelling requirement, and aim to work with smaller validation sets as well. 
 \begin{table}
 \begin{center}
 \begin{tabular}{||c|c|c|c||} 
 \hline
Work & Perspective & labeled & unlabeled\\ [0.5ex] 
 \hline\hline
 \citet{kothari2020semisupervised}& Vertical & 1200 & 15000\\
 \hline
 \citet{yalniz2019billionscale} & Horizontal & 1M & 100M\\
 \hline
 \citet{zhai2019s4l} & Horizontal & 12800 & 1.2M\\
\hline
 \citet{xie2019unsupervised} & Horizontal & 250 & 25000(S)\*\\
 \hline
 This work & Vertical &  200 & 60M\\
 \hline

\end{tabular}
\caption{Dataset details for semisupervised learning work. (S) indicates synthetic/augmented data generation as the main source of data for semisupervised learning. Perspective refers to the capture angle with vertical perspective corresponding to overhead imagery.}
\label{Tab:Semisup_compare}
\end{center}
 \end{table}


\subsubsection{Active learning}

In machine learning, active learning refers to the class of techniques where the model can iteratively query a human user regarding the ground truth of a subset of input data. Based on the user's input, the model then performs additional learning to improve its prediction accuracy. This requires manual intervention at each iteration of the learning process. Active learning has been successfully used for a multitude of tasks including crystal structure prediction \cite{podryabinkin2019accelerating}, vehicle detection \cite{sivaraman2014active} and facial recognition \cite{hewitt2006active}. These approaches work well in theory, by using an oracle or already annotated dataset for evaluation purposes. However, \citet{settles2011theories} argues that, when attempting to bootstrap a new dataset in practice, it is generally not time efficient to wait for model training to finish before annotating more images. A key difference between semi-supervised learning and active learning is that the agent doing the annotation in semi-supervised learning is an automated model, whereas in active-learning it is generally a human.


\subsection{Applications of overhead imagery}

Overhead (satellite and aerial) imagery has been used in previous research for a wide variety of applications. The features of the urban fabric provide important pointers to explore pressing issues in contemporary society. For example, information extracted from high-resolution satellite imagery has been used to estimate poverty in African countries \cite{jean2016combining} and provide disaster and crisis-management support \cite{voigt2007satellite}. Further, it has proven valuable for inferring population size \cite{robinson2017deep}, assessing land cover changes \cite{yang2002using}, and monitoring food security through agricultural crop mapping \cite{shelestov2017exploring}. Beyond imagery, satellite remote sensing has enabled global analyses of air pollution \cite{martin2008satellite}, vegetation changes \cite{meeragandhi2015ndvi}, and economic activity using night‐time lights as a proxy indicator \cite{elvidge2007nightsat}.

\begin{figure}[h]
\centering
	\subfloat[]{\setlength{\fboxsep}{0pt}\includegraphics[trim = 0mm 0mm 0mm 0mm, clip, height=4cm]{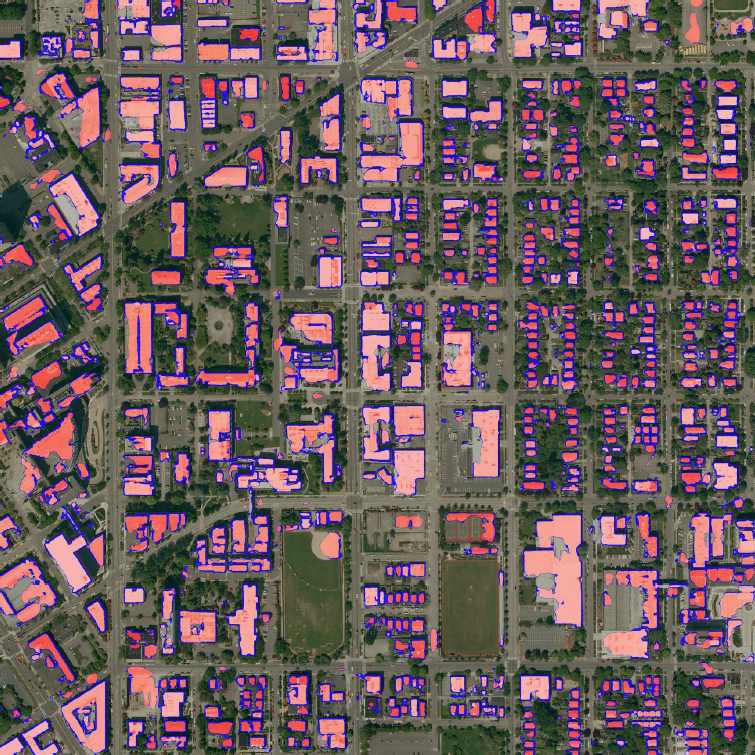}}\,
	\subfloat[]{\setlength{\fboxsep}{0pt}\includegraphics[trim = 0mm 0mm 45mm 0mm, clip, height=4cm]{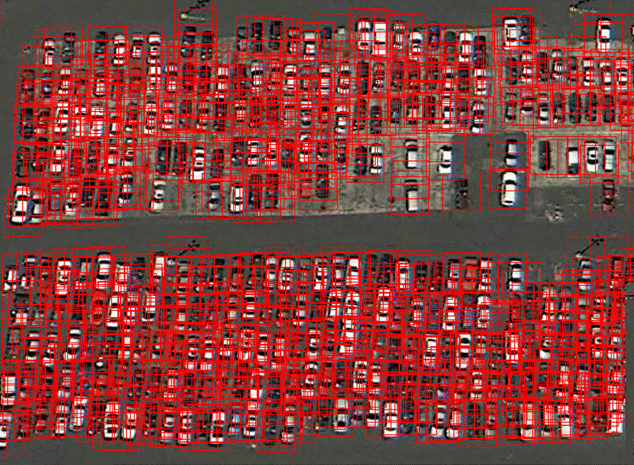}}\\
	\subfloat[]{\setlength{\fboxsep}{0pt}\includegraphics[trim = 0mm 0mm 0mm 0mm, clip, height=4cm]{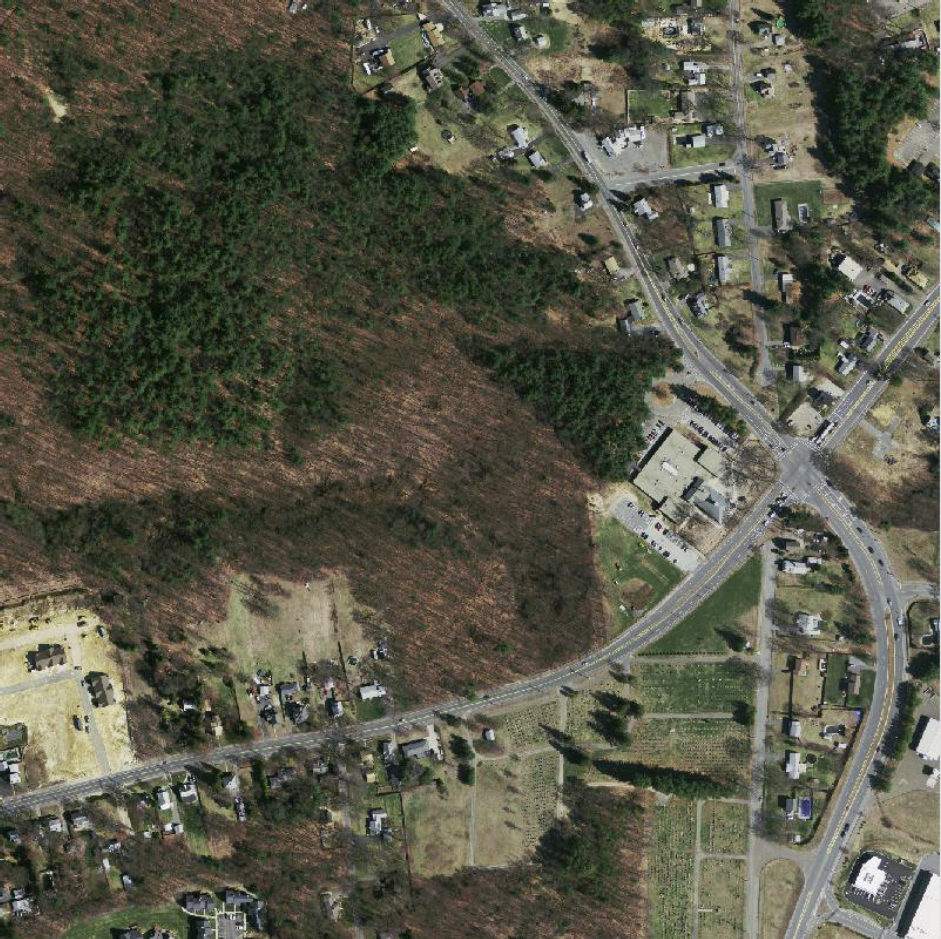}}\,
    \subfloat[]{\setlength{\fboxsep}{0pt}\includegraphics[trim = 0mm 0mm 0mm 0mm, clip, height=4cm]{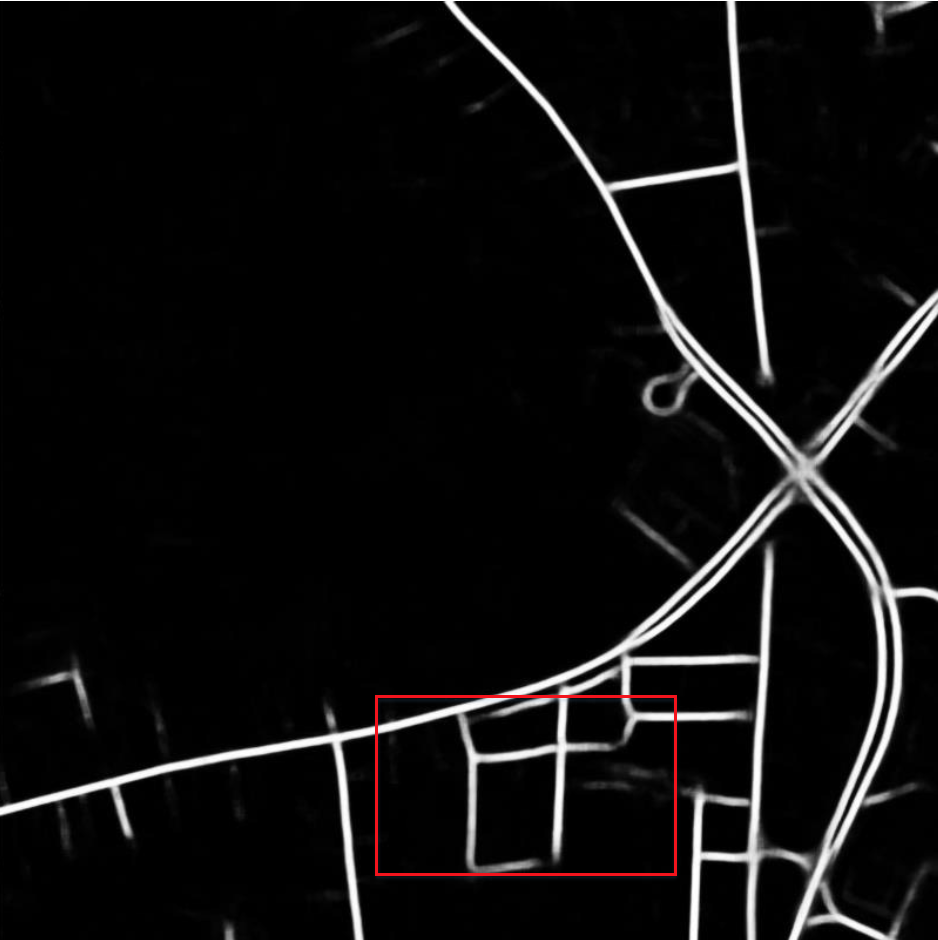}}
	\caption{Specialised object detection models: (a) buildings \cite{yuan2016automatic}, (b) vehicles \cite{chen2014vehicle}, (c--d) roads \cite{zhang2018road}}
	\label{fig:object_detection}
\end{figure}

The studies above provide evidence of the significant potential of space-based observations to explore and understand the effect of contemporary social issues on spatial organisation. While some studies implicitly use features in satellite imagery to find associative evidence, other research has focused purely on feature extraction from imagery. Importantly, the primary task of feature detection could lead to a detailed understanding of environment characteristics and enhance the explainability of research findings. In this case, the task can be formulated as an object detection problem. This research direction has been taken by various studies, generally specialised for the detection of a single object category visible in satellite imagery. For example, \citet{vakalopoulou2015building} and \citet{yuan2016automatic} developed algorithms for building detection. Further, many studies have explored the extraction of road networks from satellite imagery \cite[e.g.,][]{wang2015road, zhang2018road, mnih2010learning}. \citet{wang2015road} achieved this by predicting the road direction in satellite images and constructing the network by analysing imagery at adjacent locations. \citet{zhang2018road} created an image segmentation approach based on U-Net \cite{ronneberger2015unet} to extract road networks. More detailed characteristics of the road network can also be detected, such as specific intersection designs \cite{wijnands2020identifying}. \citet{cadamuro2019street} assessed road quality from satellite imagery, using a combination of an autoencoder \cite{ballard1987modular} and Long Short-Term Memory neural networks \cite{hochreiter1997long} to extract and analyse features. Further, \citet{chen2014vehicle} designed a methodology that can be used to detect the number of vehicles on roads. An illustration of some of these approaches is provided in Fig.~\ref{fig:object_detection}.


\subsection{Objective}
Over the past decades, technical advances in satellite remote sensing have greatly improved the quality of satellite imagery. Further improvements in image resolution have been achieved through aerial photography using airplanes, resulting in an increased availability of very high-resolution overhead imagery datasets. The additional details in high-definition aerial imagery provide opportunities for improving the accuracy of object detection methods. Further, it allows for the detection of new object classes previously undetectable from satellite imagery and difficult to collect otherwise. For example, uncommon types of infrastructure (such as cycling infrastructure) are poorly represented or incomplete in existing datasets, but can be analyzed using aerial imagery. Besides taking advantage of improvements in input data, this article explores new methods for object detection. As described above, current object detection methods are either highly specialised towards extracting a single characteristic from the environment (e.g., buildings or vehicles), or detect many classes at once with extensive manual annotation requirements. Therefore, the gap addressed in our research is the lack of a resource-efficient, generic method that can extract a more complete set of features to describe the environment in a single image. As indicated by \citet{mnih2010learning}, pre-training using unsupervised learning methods can improve model accuracy substantially, providing opportunities to develop such a generic approach.

The main motivation of this work is to enable an extensible pipeline that simplifies the collection of data for the purpose of predictive analysis across different infrastructure classes in a scalable manner. Wherever possible, the pipeline was optimized with the following objectives in mind:

\begin{itemize}
    \item Minimize human annotation effort.
    \item Flexibility to easily add more classes .
\end{itemize}

\section{Methodology}
\label{S:2}

While existing methods have explicitly explored many road related infrastructure analyses, cycling infrastructure has been poorly explored using aerial imagery. Additionally, cycling infrastructure is often clearly demarcated using specialized symbols and colorful lanes which enables its use as a well-defined type of infrastructure to explore initially using an aerial imagery analysis workflow. Therefore, while initial analysis was carried out on such infrastructure, several additional types of infrastructure and urban features were also explored to highlight the generalisability of the developed pipeline.

\subsection{Data collection}
\subsubsection{Aerial imagery}
\label{AerialImagery}

A large image dataset was obtained using an aerial imagery provider. The image collection spanned 15 of the most populated cities across Australia for a total of 62.5 million images of size 256$\times$256 pixels taken at a zoom level of 21 (corresponding to 0.074 metres per pixel at the equator, a tile edge size of roughly 20 metres and an area of roughly 400 square metres covered by each tile). The total area covered by the study was $22,536km^2$.
Data collection by city/state is indicated in Table~\ref{imagery_details}.

 \begin{table}
 \begin{center}
 \begin{tabular}{||c|c|c||} 
 \hline
City & Images & Area(km$^2$)\\ [0.5ex] 
 \hline\hline
 Melbourne & 9,018,518 & 3249\\
 
 \hline
 Sydney  & 6,700,303 & 2414\\
 \hline
 Perth  & 13,205,906 & 4758\\
 \hline
 Canberra & 4,856,845 & 1750\\
 \hline
 Adelaide & 1,286,671 & 464\\
 \hline
 Brisbane & 12,884,242 & 4642\\
 \hline
 Geelong & 4,601,846 & 1658\\
 \hline
 Bendigo & 2,112,860	& 761 \\
 \hline
 Darwin & 392,492	 & 141 \\
 \hline
 Ballarat & 3,017,364 & 1087 \\
 \hline
 Hobart & 1,043,840	 & 376\\
 \hline
 Townsville & 948,061	 & 342\\
 \hline
 Cairns & 822,028		 & 296\\
 \hline
 Wollongong & 781,521	 & 282\\
 \hline
 Toowoomba & 876,648	 & 316 \\
 \hline
 \hline
 Total &  62,549,145 & 22,536\\
 \hline
	
\end{tabular}
\caption{Imagery details}
\label{imagery_details}
\end{center}
 \end{table}


\subsubsection{Cycling infrastructure}


For the exploration of cycling infrastructure in urban environments, an initial sample of labeled imagery was obtained through an observational study \cite{meuleners2019safer}. A total of 100 participants were recruited between March 2015 and January 2017 near on-road locations in Western Australia where a crash was observed previously. Cyclists were intercepted as they stopped at traffic lights and offered a slap-band for their wrist which had the study website recruitment address printed on it. Cyclists then completed an online questionnaire and were asked to leave their contact details within the questionnaire if they were willing to be contacted to be part of the study. Potential participants were contacted by phone to further explain the study and were sent a consent form by email. Cyclists were  eligible to participate if they had not been involved as a cyclist in a bicycle crash requiring hospitalisation in the previous three years, were 18 years or older, lived in the Greater Perth area, spoke English, and cycled at least once per week. If the cyclist agreed to participate, an appointment was made to attach the GPS tracking sensors to their bicycle. Data collection included the recording of up to six hours of cycling video footage and associated GPS data per participant. Participants were asked to record any cycling they participated in and ride exactly as they normally would.

\begin{figure}[h]
\centering
    \includegraphics[angle=0,scale=0.50]{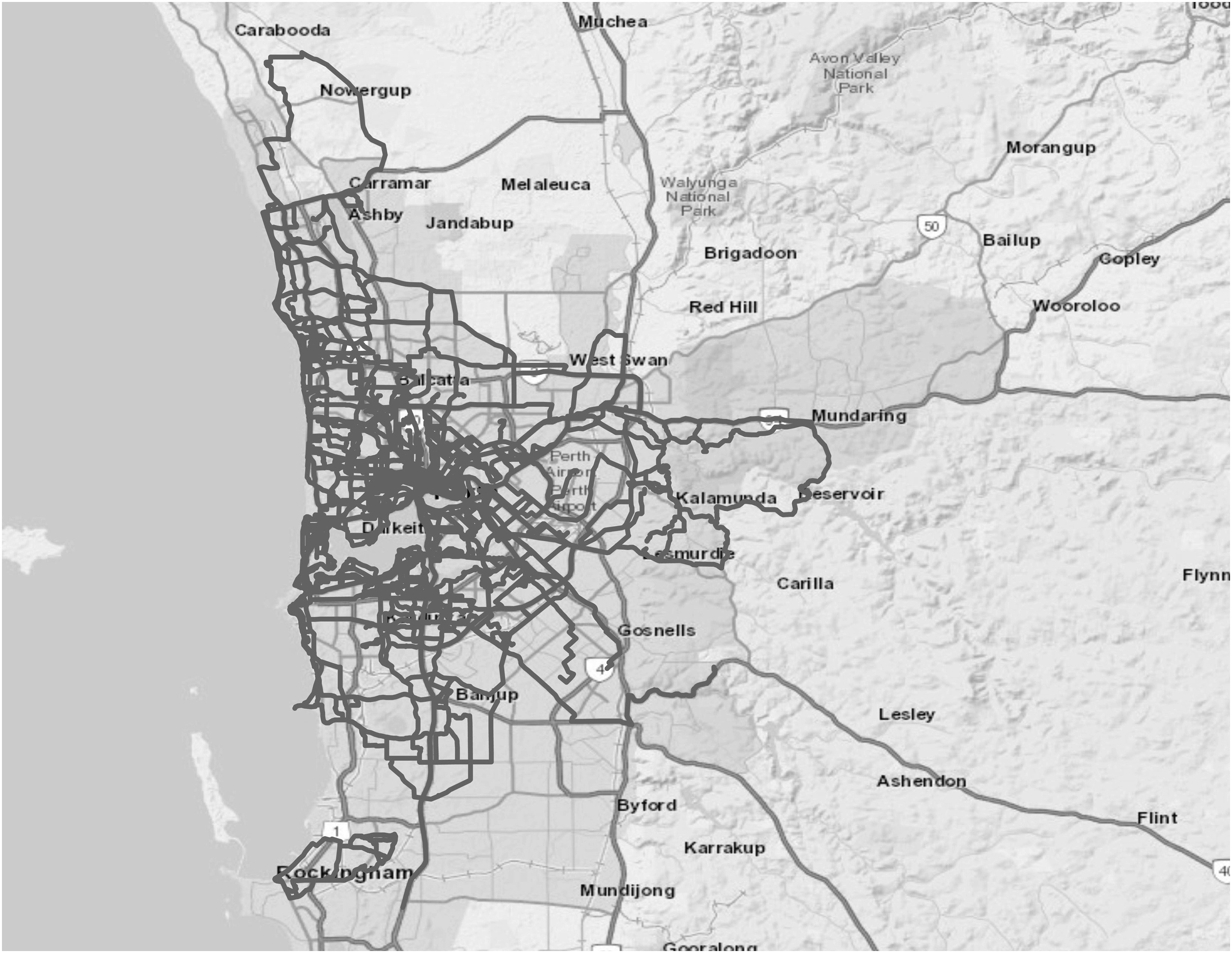}
    \caption{Cycling network extracted from GPS traces near Perth, Western Australia}
    \label{fig:WA_GPS}
\end{figure}

After the conclusion of the observation period, the recorded GPS information was allocated to specific participant trips. The most common routes that each participant travelled on covered a total road network of 1680 kilometres in Western Australia (see Fig.~\ref{fig:WA_GPS}). This consisted of 280 kilometres of bicycle paths and 1400 kilometres of on-road routes. The recorded GPS tracks were then used to annotate aerial imagery in Perth with the presence of cycling infrastructure, leading to an initial dataset of labeled imagery.

\subsection{Self-supervised representation learning}

As discussed in \ref{S2:selfsupervised}, self-supervised learning techniques allow the use of an unlabeled dataset to build a task-independent representation of the images in the dataset. This representation can then be used for other downstream tasks. In this work, our motivations for the use of self-supervised learning techniques are that they:

\begin{itemize}
    \item scale well in terms of predictive accuracy with datasets where a large portion of the data is unlabeled.
    \item allow for the rapid creation of classifiers by either transfer learning or by building a single layer on top of the existing representation.
    \item allow a single learned representation to be reused across multiple infrastructure identification tasks, allowing a considerable amount of computational work to become front-loaded and one-off.
\end{itemize}


An experiment was carried out to evaluate the suitability of such techniques, which are traditionally used on images taken from a horizontal perspective, for use with overhead imagery and map imagery. 

As an initial selection step, SimCLR \cite{chen2020simple} and Momentum Contrast (MoCo) \cite{he2020momentum, chen2020improved} were evaluated alongside a convolutional autoencoder (AE). Evaluation was carried out for the city prediction task in \cite{thompson2020global} using satellite image data for 200 cities. MoCo (95\%) had the highest validation accuracy, while SimCLR (24\%) and AE (20\%) performed significantly worse. Utilising the large batch size for SimCLR reported in the original paper (8192) for building the self-supervised representation was problematic due to computing resource limitations in terms of GPU memory. Instead, a much smaller batch size (64) had to be utilized for evaluation purposes. The original paper discusses the representation learning batch size as an important parameter for learning a general representation, as it impacts the difficulty of the pretext task used for self-supervised learning. Since MoCo provided considerably better results with a manageable batch size (256) and has been previously successfully used with remote sensing imagery\cite{seneviratne2021contrastive}, MoCo was selected for future experimental work.

For validating the utility of MoCo further for this use case, we refer to \cite{seneviratne2021Selfsup}. \citet{seneviratne2021Selfsup} conducted an experiment to verify the applicability of MoCo and to identify the scalability of this method to unseen classes (cities). The city prediction task discussed previously was used, but while representation learning (the pretraining step) was carried out on either 200 or 1667 cities, model training and testing was carried out under two settings: 200 cities and 1667 cities. For the 200 cities, the same 200 cities as in pretraining were used: checking for the ability for the representation to cover tasks or classes captured within the pretraining data itself. With pretraining on 200 cities and training/evaluating on 1667 cities, consistency of the model in representing both previously seen classes and unseen classes was evaluated. This result is important as the class-independent or generic nature of the representation would be crucial for allowing reusability across other problem domains with multiple classes (such as different types of infrastructure). Pretraining and training was carried out on the ResNet50 architecture with a batch size of 256. For training, a high learning rate of 30 was used with stochastic gradient descent since only a single layer needed to be trained (matching a standard workflow employed in previous self-supervision based studies) \cite{he2020momentum,seneviratne2021contrastive}. Detailed results in Table ~\ref{tab:selfsup_dicta} indicate significant potential for the use of self-supervision to extend to new classes previously unseen by the pre-trained representation.

 \begin{table}
 \begin{center}
 \begin{tabular}{||c|c|c|c|c|c||} 
 \hline
 Imagery & V1/V2 & Pretrain cities& Pretrain epochs & test cities&  Acc \\ [0.5ex] 
 \hline\hline
 Satellite& V1 & 200 & 200 & 200 & 95\% \\ 
 \hline
 Satellite& V2 & 200 & 200 & 200 & 99\% \\ 
 \hline
 Satellite& V1 & 200 & 200 & 1667 & 81\% \\ 
 \hline
 Satellite& V2 & 200 & 200 & 1667 & 95\% \\ 
 \hline
 
 Satellite& V2 & 1667 & 145 & 1667 & 98\% \\ 
 \hline
 Maps& V1 &1667 & 200 & 1667 & 67\% \\
 \hline
 Maps& V2 &1667 & 200 & 1667 & 61\% \\
 \hline
\end{tabular}
\end{center}
\caption{
Testing on 1667 cities with 1000 images per city with an 80\%/20\% training/validation split on the city prediction task.
V1/V2 refers to the version of MoCo used.}
\label{tab:selfsup_dicta}
\end{table}


\subsubsection{Ablation on using self-supervision}
\label{Ablation-selfsup}
An ablation test was performed on the above workflow, for validating its usefulness with the aerial imagery. This was achieved by sampling 100 images each for training and 1000 images each for validation for two classes (cycle infrastructure vs other) from the aerial dataset mentioned in Section \ref{AerialImagery}. A ResNet50 model was then built and trained on this task following three separate configurations. The first was instantiated with the pre-trained weights from ImageNet\cite{deng2009imagenet} for ResNet50, which is a commonly used approach in computer vision. An objective of this experiment is to evaluate the suitability of such a technique for use with overhead imagery. A single fully connected layer was trained for the purposes of class prediction, and was placed on top of the final bottleneck layer of the ResNet network (identical to ImageNet training except for the number of classes). The second configuration used a pretrained representation built from 100,000 unlabeled images from the aerial imagery dataset. These pretrained weights were used instead of the weights loaded from the pretrained ImageNet model. For both these configurations, the layers of the ResNet are frozen and the corresponding weights are not updated during training. This ensures that the model is forced to rely on only its pretrained representation as a feature extractor, while learning only very high level abstract concepts relating to the task at hand. A high learning rate of 30 was used with stochastic gradient descent since only a single linear layer was to be trained. The third configuration uses the pretrained representation learned in the second configuration, but uses it for end-to-end transfer learning. In this configuration, all the weights of the ResNet are updated during the training process, which is not the case in the other configurations. A learning rate of 0.001 was used with stochastic gradient descent in order to minimize changes to the pre-trained weights under this configuration. This low weight aims to minimize the destruction of pre-learned features in the model, by only performing small tweaks instead of big shifts in existing features. The neural network was trained for 200 epochs and the checkpoint with the best validation performance was used for reporting performance. The results are in Table~\ref{ablation_results} under Section~\ref{results:ablation}.



\subsubsection{Characterizing self-supervised performance}
\label{experiment:selfsup_characterization}
As an initial evaluation of transfer learning from the self-supervised representation, an experiment was carried out to evaluate ResNet50 based on finetuning by transfer learning from the frozen MoCo representation. Results are in Table~\ref{transfer_initial}. The objective of this experiment was to better characterize performance of the two configurations built on the pretraining workflow. The complete dataset of training set, validation set and test set each representing 2 classes, contained 33,337 images. These aerial images were randomly selected from a large set of labeled road images sampled from areas known to have cycling infrastructure. Images containing cycling infrastructure were manually filtered such that 18,642 images contained cycling infrastructure, and 14,695 did not contain any cycling infrastructure. The ResNet50 architecture was used in all experimentation and the highest validation accuracy model was picked as the final model. For the \say{Frozen} configuration, a learning rate of 30 and a batch size of 4 was used with stochastic gradient descent, while for the \say{Transfer} configuration, a learning rate of 0.001 was used alongside a batch size of 16 with stochastic gradient descent. By testing different configurations of training and validation set sizes, the expectation was that a better understanding of model performance scaling with larger training set sizes could be obtained. This would in turn serve to confirm the results in Table~\ref{ablation_results} while indicating potential thresholds in terms of manual annotation requirements for solving tasks of this nature. The overall size of the dataset is kept fixed to more accurately mirror the actual situation of using a model to iteratively grow a dataset from a pool of unlabeled images: the size of the unlabeled image set would shrink as more images are moved out of the unlabeled dataset. The results are in Table~\ref{transfer_initial} under Section~\ref{results:selfsup_characterization}.





%

\subsection{Semi-supervised learning}

Semi-supervised learning was explored as a means of generating more accurate models as well as for creating a workflow capable of utilizing the large dataset available to its maximum potential. 

There are two main configurations used in this regard, with training details broadly in line with previous experiments: Frozen and Transfer. The main focus of this section is exploring techniques that allow the training set of the model workflow to continually expand, thereby creating more accurate models. This creates a positive feedback loop that can be used with minimal manual tuning to automatically label and process the entire dataset.

\aside{
\begin{itemize}
    \item Frozen configuration - which builds upon an initial representation and simply trains a linear fully connected layer on top, while freezing the initial representation.
    \item Transfer configuration - which performs fine-tuning/transfer learning on the initial representation in order to tweak the representation to further align it with the task being trained on.
\end{itemize}
}

\subsubsection{Initial semi-supervised experiment}
\label{single_class_bootstrap_eval}

To evaluate the suitability of semi-supervised learning, an experiment was carried out using the above configurations. These configurations were evaluated on a single task (cycling infrastructure categorization) on the same dataset of 33,337 images as in Section\ref{experiment:selfsup_characterization}. The results can be found in Table~\ref{bootstrap_1_cycle} under Section~\ref{Results:Semisupervised_Initial}.

\subsubsection{Semi-supervised consistency}
\label{Experiment:SemiSupervised_Consistency}

As a follow-up experiment, the consistency of continued semi-supervised learning was explored as a single-class fixed dataset experiment. 
A priority queue based implementation was used to track the top 500 highest and lowest cycle symbol confidence predictions from the test set to merge into the training set. The validation set was fixed at 1000 images each. Continuous evaluation of the bootstrapping approach using the transfer learning from the Frozen configuration was carried out, starting with 1000 training and validation images per class with a step size of 500. The results are in Table~\ref{bootstrap_multiple_cycle} under Section~\ref{Results:SemiSupervised_Consistency}
.




\subsubsection{Analysis of multiple classes using Frozen configuration}
\label{linearlayer}

While previous experiments were exclusively single class (looking at cycle symbol classification), this experiment aims to evaluate the methodology in a more generic manner. A practical limitation in this regard is the image annotation requirement for attempting many different tasks. To make the most of limited annotator time, a limit of 200 annotations per class per task was imposed, with 100 images each for the training and validation sets respectively. As before, the two classes correspond to a ``Task" class vs a "Background" class. The main reason for this experimental setup is that it is highly likely for multiple infrastructure classes to be present in the same image. Therefore, by creating a binary classification task, we are able to overlap annotations from multiple models on the same image in a similar fashion to object detectors, without needing to generate bounding boxes for the different tasks which would severely limit annotator time availability to explore multiple classes. Within this limitation of 100 training images, prior experiments (Section~\ref{Ablation-selfsup}) indicate that the Frozen configuration performs best and that a validation set of 100 images should be sufficient in this respect. Evaluation of the Frozen configuration trained on 100 images of each class in training and validation is provided to compare the base performance of the methodology on each task. The Frozen configuration is used as it is useful for providing a baseline level of performance to compare against. Further, it has the added benefit of being trained very fast due to the high learning rate used. The percentages reported correspond to the precision of the class under investigation. No false negatives were detected during this set of experiments. Evaluation was carried out on a random sample of 100 images drawn uniformly from the top 1000 predictions at each location ordered by confidence. The results of this experiment can be found under Section~\ref{results:frozen}.


\subsubsection{Automated analysis using archival semi-supervised learning}
\label{sec:Archival_Initial}
This experiment explored the development of a workflow centered around using historic images at locations for improving model accuracy. In particular, the main objective was to build upon the results from Section~\ref{linearlayer} by using historical imagery as a data augmentation/semi-supervised learning strategy. 

To this end, several key semantics of the task at hand are exploited. The key insight for this methodology is that infrastructure is static: if it is available at a location at present, it is likely to have been present at that location in the recent past. It is also reasonable to expect that the probability of finding that infrastructure would decrease if the image was taken at an earlier date than a later date, simply because the infrastructure might have been constructed at an intermediate date. By contrast for the background class: if a particular image does not contain some infrastructure it is highly unlikely to have been there in the past: effective planning schemes mean that cities and other infrastructure are usually planned well ahead of time, and drastic changes are unusual in the short term.

Hence, the following assumptions are made regarding historical images at a location:
\begin{itemize}
    \item The probability of finding the task class in historical images at a location correctly labeled as the background class is negligible.
    \item The probability of finding the task class in historical images at a location correctly labeled as the task class is high, with the probability of finding the task class in more recent images being higher than in older images.
\end{itemize}

Considering the model as a \say{task} class detector, a false positive confounder would be an image of the \say{background} class being incorrectly classified as belonging to the \say{task} class (identical to a false positive). Let $\Phi$ be the class of all historical images at all background image locations from the training set. Then, consider the set  $\Theta \subset \Phi$ which contains all the confounders from the model performing inference on images contained in $\Phi$. The set $\Theta$ is then a very informative dataset for the current model to learn from, as the model has been unable to classify them correctly, despite having seen a preceding image in the training set in the background class. Additionally, more recent confounders would be more useful than older ones, as the more recent images could be expected to look more structurally similar to the image at present and therefore contain more interesting features to include in the background class (as opposed to, for example, an unbuilt area from a long time ago which would likely not add much predictive value to the background class). Note that this logic is not necessarily commutative if the \say{task} and \say{background} classes are swapped: the infrastructure under investigation might have been constructed/painted very recently and thus, may not necessarily be misclassified as being \say{background} class (since if the \say{task} class is not present in the image, it belongs, by definition, to the \say{background} class). Conceptually, this is similar to boosting\cite{schapire2003boosting} in machine learning, as images misclassified by the model are assigned with an increased weight into the training set thus increasing their importance in terms of contribution to the decision boundary of the model.

Let $\Phi_T$ be the set of all historical images corresponding to training set locations and $\Phi_B$ be the set of all historical images corresponding to background locations, with $\Phi = \Phi_T \cup \Phi_B$. Note that the latest available images also count as historical images by definition and as such would be included in these sets. Due to being historical images of labeled locations, clearly the sets $\Phi_T, \Phi_B$ contain images that the model could learn from, in a supervised manner. However, not all images would be equally useful or correct to learn from. Thus, assigning a weight to each individual historical image allows the training process to be controlled (when assigned a weight of zero, an image would essentially have no impact on the training process). Therefore, the problem at hand can be defined as follows:

Let $\Phi_T^i$ and $\Phi_B^j$ correspond to the above sets with $i,j$ corresponding to arbitrary orderings (indices). Then let the individual loss of each training sample be determined by the function L(x), which would apply the loss function used in the neural network to the corresponding output of x. Then, the overall loss function becomes:

\begin{equation}
    Loss = \sum_{i}{\alpha_T^iL(\Phi_T^i)} + \sum_{j}{\alpha_B^jL(\Phi_B^j)}
    \label{Eqn:Loss}
\end{equation}

Where $\alpha_T^i \in \mathbb{N}$ corresponds to individual historical task weights and $\alpha_B^j \in \mathbb{N}$ corresponds to individual historical background weights. Without loss of generality and for simplicity, let the first N elements of both orderings $\Phi_T^i$ and $\Phi_B^j$ be set to an arbitrary ordering of the initial human labeled training set of N images per class. As the model trains in a semi-supervised manner, the main difference in the data composition is tracked by the different values of $\alpha$ over the entire dataset. Note that images with $\alpha = 0$ have no contribution to model training, and may be omitted during training. 


The following operations are defined in order to modularise the workflow for the semi-supervised learning process for improving the performance of the models using archival imagery. It is important to note that confidence metrics are defined with respect to the \say{task} class. The confidence metric corresponds to the probability of a particular image belonging to the \say{task} class and is related to the probability of the image belonging to the background class as $P_T = 1 - P_B$ due to the presence of only 2 classes.

\begin{itemize}
\item \textbf{Train} - Builds a classifier from the currently available training dataset as defined by Equation~\ref{Eqn:Loss}.
\item \textbf{Predict} - Uses the most recently built classifier to perform prediction on the historical datasets ($\Phi_T$ and $\Phi_B$ separately) and assigns confidence scores based on the \textbf{task} class (not the background class).

The semi-supervised learning process relies only upon training several iterations of computer vision models which have access to different training sets. The \textbf{train} and \textbf{predict} operations provide interfaces for this functionality. As any modifications to the data/weights only affect the process once a model is trained and prediction is carried out on $\Phi_T$ and $\Phi_B$ (thus updating the confidence metrics), each step/iteration of the semi-supervised learning process begins with training the model and predicting on $\Phi_T$ and $\Phi_B$.

\item \textbf{Update Task} - Increments $\alpha_T^i$ corresponding to the $M_T$ highest confidence task detections in $\Phi_T$.

\item \textbf{Update Background} -  Increments $\alpha_B^j$ corresponding to the $M_B$ lowest confidence task detections in $\Phi_B$.

\item \textbf{Update Confounders} - Increments $\alpha_B^j$ corresponding to the $M_C$ highest confidence task detections in  $\Phi_B$ (hence matching the definition of confounder: high confidence, but assigned to the wrong class).


The \textbf{update} operations are used for managing the datasets over iterations of the semi-supervised learning process. By updating the contribution of each individual image to the loss function, the decision boundary of the model is modified as well, with some images receiving a higher importance than others. It is important to note that order statistics (such as the $M_T$-th largest confidence value) need to be maintained independently for the two datasets $\Phi_T$ and $\Phi_B$ as the underlying semantics and class probability distributions for the two datasets are very different.
\end{itemize}

In combination, these operations define the behaviour of the semi-supervised technique. The temporal control of image availability over time is managed by gradually broadening the time frame over which the model is allowed to update values of $\alpha$: initially, only values corresponding to more recent images may be updated but in later iterations, $\alpha$ values corresponding to earlier images may be updated as well. This greatly decreases the probability that the model will incorrectly classify an image due to having less of a connection (structurally or otherwise) to the image's present predecessor image. This behaviour is determined by the parameters $D_T$ and $D_B$ corresponding to the \say{task} and \say{background} classes and denotes the maximum duration (in months) from the latest image in the dataset that an image would need to have been captured, in order for the $\alpha$ value to be updatable. In other words, $\alpha_T^i$ may be updated if and only if the image $\Phi_T^i$ was captured within $D_T$ months of the last image in $\Phi_T$, and similarly for $\alpha_B^j,D_B,\Phi_T^i$ and $\Phi_T$.

\hfill

\begin{algorithm}[H]
\SetAlgoLined
  
  \textbf{Signature:} Update($\Phi[1...N], \alpha[1...N], conf[1...N], D, K, top, date_{ref})$ \\
    $\Phi$ - dataset of N images corresponding to $\alpha$ values\\
    $\alpha$ - array of individual image loss contributions\\
    conf - array of confidence values corresponding to dataset\\
    D - duration corresponding to dataset $\Phi$\\
    K - number of $\alpha$ values to be modified\\
    top - Boolean indicating if top or bottom K alpha should be updated\\
    date$_{ref}$ - reference date for duration comparison\\
    \hfill \\
  \textbf{Execution:}
  
  \uIf{top == True}{
   \textbf{\#} Select k-th largest confidence\\
   conf$_K$ = QuickSelect(conf, N-K)\\
   }\Else{
   \textbf{\#} Select k-th smallest confidence\\
   conf$_K$ = QuickSelect(conf, K)\\
  }   
  
  \While{$0 \leq i < N$}{
    \uIf{$top == True$ \textbf{and} $conf[i] > conf_K$ \textbf{and}  $(\Phi[i].date - date_{ref}) < D$}{
        $alpha[i] = alpha[i] + 1$
    }
    \If{$top == False$ \textbf{and} $conf[i] < conf_K$ \textbf{and}  $(\Phi[i].date - date_{ref}) < D$}{
        $alpha[i] = alpha[i] + 1$
    }
  }

 \label{Alg:update}
 \caption{Update algorithm}

\end{algorithm}

As some of the novelty of this work is concentrated in the update operations, we provide an algorithmic implementation of the base update functionality in Algorithm~\ref{Alg:update}. Note that an implementation which uses a heap-based approach for tracking the top and bottom K-th confidence items in a given dataset with complexity O(N log K + K log K) for maintaining and iterating over such a list of items. An alternate implementation is to use the QuickSelect\cite{mahmoud1995analysis} algorithm to generate the K-th order statistics for a given unsorted dataset. This allows the creation of a list of the top or bottom K confidence items in O(N + K) time and is presented in Algorithm~\ref{Alg:update}. Since the discussed applications are not time-critical and because K is generally much smaller than N in most situations, both implementations are expected to have similar performance and are listed here for the sake of completion. The update operations previously mentioned are implemented in Algorithm~\ref{Alg:update_all}.

\SetKwIF{If}{ElseIf}{Else}{if}{}{else if}{else}{end if}%

\begin{algorithm}[H]
\SetAlgoLined
\textbf{Signature}: UpdateTask($\Phi_T$[1...N], $\alpha_T$[1...N],conf$_T$[1...N],D$_T$,M$_T$,date$_{ref})$ \\
\textbf{Execution}:\\
  Update($\Phi_T$, $\alpha_T$,conf$_T$,D$_T$,M$_T$,top=True,date$_{ref})$ \\
  
\hrulefill\\
\textbf{Signature}: UpdateBackground($\Phi_B$[1...N], $\alpha_B$[1...N],conf$_B$[1...N],D$_B$,M$_B$,date$_{ref})$ \\
\textbf{Execution}:\\
  Update($\Phi_B$, $\alpha_B$,conf$_B$,D$_B$,M$_B$,top=False,date$_{ref})$ \\
 
\hrulefill\\
\textbf{Signature}: UpdateConfounder($\Phi_B$[1...N], $\alpha_B$[1...N],conf$_B$[1...N],D$_B$,M$_C$,date$_{ref})$ \\
\textbf{Execution}:\\
  Update($\Phi_B$, $\alpha_B$,conf$_B$,D$_B$,M$_C$,top=True,date$_{ref})$ \\
  
  
   
 \label{Alg:update_all}
 \caption{Updates for Task,Background and Confounder}
\end{algorithm}

Algorithm~\ref{Alg:overall} contains the overall procedure (corresponding to a single step from Table~\ref{Tab:operations_overview}), building upon Algorithms~\ref{Alg:update} and~\ref{Alg:update_all}. Additional subscripts are provided to indicate the recommended parameters for the function based on the previously defined data arrays.

\begin{algorithm}[H]
\SetAlgoLined
\textbf{Signature}: \\SSMS($\Phi$[1...N], $\alpha[1...N]$, conf[1...N], D$_T$, D$_B$, M$_T$, M$_B$, M$_C$, date$_{ref})$ \\
\textbf{Execution}:\\
\#Train the model based on equation~\ref{Eqn:Loss}\\
model = Train($\Phi$, $\alpha$)\\
\#Update the predictions based on the new model\\
conf = Predict($\Phi$)

  UpdateTask($\Phi_T$, $\alpha_T$,conf$_T$,D$_T$,M$_T$,date$_{ref})$ \\
  UpdateBackground($\Phi_B$, $\alpha_B$,conf$_B$,D$_B$,M$_B$,date$_{ref})$ \\
  UpdateConfounder($\Phi_B$, $\alpha_B$,conf$_B$,D$_B$,M$_C$,date$_{ref})$ \\
 \label{Alg:overall}
 \caption{Semi-supervised model step(SSMS)}
\end{algorithm}

\newcommand{\comment}[1]{}

\comment{
TODO: probably remove below.
Overall process is in Table~\ref{Tab:operations_overview}.

\begin{table}
 \begin{center}
 \begin{tabular}{||c|c|c||} 
 \hline
Operation & Task & Background\\ [0.5ex] 
 \hline\hline
 Train and Predict & 200 & 200\\
 \hline
 \textbf{Expand} & 2020-06 & 2019-01\\
  \hline
 Confounder Merge 500 & 200 & 700(+500)\\
 \hline
 Train and Predict & 200 & 700\\
 \hline
 Merge Task 50 & 250(+50) & 700\\
 \hline
 Confounder Merge 500 & 250 & 1200(+500)\\
 \hline
 Train and Predict & 250 & 1200\\
 \hline
 \textbf{Expand} & 2020-01 & 2019-01\\
  \hline
 Merge Task 100 & 350(+100) & 1200\\
 \hline
 Confounder Merge 500 & 350 & 1700(+500)\\
 \hline
 Train and Predict  & 350 & 1700\\
 \hline
 Merge Task 150  & 500(+150) & 1700\\
 \hline
 Confounder Merge 500  & 500 & 2200(+500)\\
 \hline
 Train and Predict  & 500 & 2200\\
 \hline
 \textbf{Expand} &  2019-01 &  2014-01\\
  \hline
 Merge Task 150  & 650(+150) & 2200\\
 \hline
 Merge Background 250  & 650 & 2450(+250)\\
 \hline
 \textbf{Expand} &  2017-01 &  2014-01\\
  \hline
 Merge Task 350  & 1000(+350) & 2450\\
 \hline
 Merge Background 350  & 1000 & 2800(+350)\\
 \hline
 Merge Background ALL  & 1000 & 2800 + ALL\\
 \hline
\end{tabular}
\end{center}
\caption{Process of semi-supervised learning}
\label{Tab:operations_overview}
\end{table}

}

\begin{table}
 \begin{center}
 \begin{tabular}{||c|c|c|c|c|c|c|c||} 
 \hline
Step & $D_T$ & $D_B$ & $M_T$ & $M_B$ & $M_C$ & $\sum_{i}{\alpha_T^i}$ & $\sum_{i}{\alpha_B^j}$\\ [0.5ex] 
 \hline\hline
 0 & 0 & 0 & 0 & 0 & 0 & 100 & 100\\
 \hline
 1 & 6 & 24 & 0 & 0 & 500 & 100 & 600\\
 \hline
 2 & 6 & 24 & 50 & 0 & 500 & 150 & 1100\\
 \hline
 3 & 12 & 24 & 100 & 0 & 500 & 250 & 1600\\
 \hline
 4 & 12 & 24 & 150 & 0 & 500 & 400 & 2100\\
 \hline
 5 & 24 & 48 & 150 & 250 & 0 & 550 & 2350\\
 \hline
 6 & 48 & 84 & 350 & 350 & 0 & 900 & 2700\\
 \hline
\end{tabular}
\end{center}
\caption{Summary of semi-supervised learning iterations}
\label{Tab:operations_overview}
\end{table}

After the steps corresponding to Algorithm~\ref{Tab:operations_overview}, all $\alpha_B^j$ are incremented by one to guarantee inclusion in the training set (justified by the assumption that the class of infrastructure being detected was never present at background image locations). The final model used for querying the unlabeled non-archival dataset is then trained using transfer learning as in previous experiments. The results from this experiment are in Section~\ref{results:Archival_semi}.

\subsubsection{Specialized Representation Learning experiment}
\label{sec:archival_specialized}
A natural follow up to the experiment in Section~\ref{sec:Archival_Initial} is to specialize the initial representation learned in a task-independent fashion. Additionally, the results from the ablation study in Section~\ref{Ablation-selfsup} indicate the benefits of specializing the representation in a more domain specific manner: ImageNet weights are often used as a generic task-independent representation, but by specializing the representation for aerial imagery, significant improvements are obtained. 

The road images in this experiment were generated as the top 100,000 confidence predictions from the overall dataset by the road model from Table~\ref{archival_intial_multitask} which had a very high precision. This pre-training set was validated by sampling 10,000 images uniformly at random and performing manual annotation. The precision over this sample was 100\%. Using this 100,000 image dataset,  a further specialized representation was created by using an identical workflow to the initial task-independent representation. With this specialization, the representation is still task-independent, however, it is no longer as generic as the previous pre-trained representation (used in section \ref{sec:Archival_Initial}) and better performance can be expected on tasks specifically associated with roads, and poorer performance otherwise. The results of this experiment are in Section~\ref{results:specialized_road}.

\subsubsection{Archival Imagery analysis}

\begin{figure}[h]
    \subfloat[]{\setlength{\fboxsep}{0pt}\includegraphics[angle=0,scale=0.32]{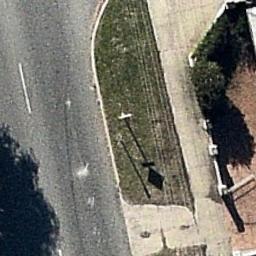}}\,
    \subfloat[]{\setlength{\fboxsep}{0pt}\includegraphics[angle=0,scale=0.32]{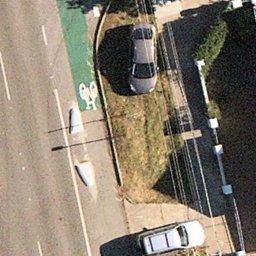}}\,
    \subfloat[]{\setlength{\fboxsep}{0pt}\includegraphics[angle=0,scale=0.32]{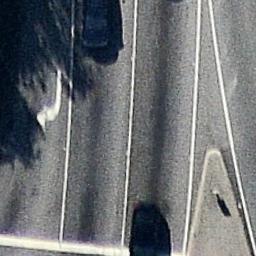}}\,
    \subfloat[]{\setlength{\fboxsep}{0pt}\includegraphics[angle=0,scale=0.32]{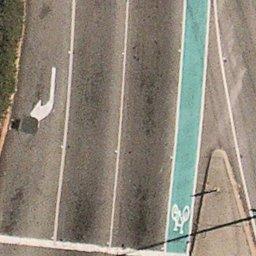}}
    \caption{Manual archival imagery evaluation at two sample locations in Melbourne, Australia. Both locations were captured in 2018 (a,c) and 2020 (b,d).}
    \label{fig:archivalEval}
\end{figure}

The results of archival analysis are in Fig.~\ref{fig:archivalEval}. From an infrastructure analysis perspective, this allows analysis of the growth of infrastructure at the city level. Since infrastructure forms a key cornerstone of cities that affects all other aspects including transport and health, being able to analyse when specific classes of infrastructure were introduced is very useful. In particular, analysis across multiple classes of infrastructure is very valuable in understanding the relationship between such classes and other aspects of cities such as inhabitant behaviour and population health outcomes. This can also be used to identify the trajectory of cities with regards to how well they are supporting healthy habits through initiatives such as provision of safe, active transport infrastructure for citizens.

From a computer vision perspective, datasets such as these introduce new ways of utilizing geographical information spanning multiple time-steps. For example, in this work we have exploited the static nature of infrastructure. However, it is possible to take this even further by exploiting the fact that once infrastructure is introduced to a location, it is very likely to stay there. This \say{expectation of the maintenance of infrastructure} allows the introduction of additional analytical steps that can improve model performance. In this situation, for example, we can expect an example vector over multiple time steps at the same location to look like (False, False, False, True, True) with the infrastructure class not being present in the first three time-steps and being introduced sometime between the third and fourth time-steps. It is reasonable, then, to assume that it will also be present thereafter. Therefore, if we assume that for some time-step $t_0$, that the infrastructure was introduced to the location between $t_0$ and $t_0+1$, then the prediction for that location would take the form of a \say{step} function, with confidence zero upto $t_0$ and confidence one beyond $t_0+1$. This could be used to optimize the infrastructure detection models either by introducing this as a consistency loss which penalizes how different the model's characteristic function is from a step function, or by enforcing such behaviour by performing smoothing operations on the confidence scores across multiple time-steps at the same location.


\section{Results}
\label{S:3}

\subsection{Self-supervised learning}
\subsubsection{Ablation on using self-supervision}
\label{results:ablation}


\begin{table}
 \begin{center}
 \begin{tabular}{||c|c|c|c|c|c||} 
 \hline
 Config & Technique & Model & Training & Validation &  Accuracy \\ [0.5ex] 
 \hline\hline
 1 & MoCoV2 Frozen & ResNet50 & 100 & 1000 & 72\% \\ 
 \hline
 2 & ImageNet Frozen & ResNet50 & 100 & 1000 & 66\% \\ 
 \hline
 3 & MoCoV2 Transfer & ResNet50 & 100 & 1000 & 61\% \\ 
 \hline
  1 & MoCoV2 Frozen & ResNet50 & 1000 & 1000 & 70\% \\ 
 \hline
 2 & ImageNet Frozen & ResNet50 & 1000 & 1000 & 70\% \\ 
 \hline
 3 & MoCoV2 Transfer & ResNet50 & 1000 & 1000 & 92\% \\ 
 \hline
\end{tabular}
\end{center}
\caption{Evaluation of the impact of different initializations and training set sizes on performance. \textbf{MoCoV2} refers to the self-supervised representation trained as part of this work. \textbf{ImageNet} refers to the standard deep learning representation of a model pretrained on the ImageNet dataset. \textbf{Frozen} refers to disabling parameter update in most of the neural network, while \textbf{Transfer} refers to allowing parameter update in most of the neural network. 1000 images were used for validation and a ResNet50 model was used as the
architecture.}
\label{ablation_results}
 \end{table}

The results (Table~\ref{ablation_results}) for the ablation on using self supervision detailed in Section~\ref{Ablation-selfsup} indicate that the representation learned by MoCo is superior. Interestingly, allowing the model to modify the representation learned by MoCo (Configuration 3) leads to a drop in holdout accuracy from 72\% (from Configuration 1) to 61\% for the smaller training set size, which is indicative of the issue of overfitting to the data.

\subsubsection{Characterizing self-supervised performance}
\label{results:selfsup_characterization}

The results in Table~\ref{transfer_initial} correspond to the validation accuracy obtained using the self-supervised representation as part of the experiment described in Section~\ref{experiment:selfsup_characterization}.

\begin{table}
 \begin{center}
 \begin{tabular}{||c|c|c|c|c|c|c|c|c|c||} 
 \hline
Method &  Train & Val & Test & TP & TN & FP &  FN & Acc(Val) & Acc(Test) \\ [0.5ex] 
 \hline\hline
 Frozen & 100 & 1000 & 31137 & 13892 & 8181 & 5334 & 3730 & 73\% & 71\% \\ 
  \hline
 Frozen & 100 & 100 & 32937 & 15071 & 7718 & 6697 & 3451 & 70\% & 69\% \\ 
 \hline
 Transfer & 1000 & 1000 & 29337 & 16015 & 10924 & 1691 & 707 & 93\% & 92\% \\ 
 \hline
  Transfer & 5000 & 1000 & 21337 & 12223 & 8301 & 314 & 499 & 96.5\% & 96.2\% \\ 
 \hline
\end{tabular}
\end{center}
\caption{Characterizing model behaviour over different training/validation configurations on a fixed size dataset. \textbf{Train,Val,Test} - number of training,validation and testing images used, respectively. \textbf{TP} = True Positives, \textbf{FN} = False Negatives. \textbf{Acc(Val)} and \textbf{Acc(Test)} correspond to accuracy on the validation and test set respectively.}
\label{transfer_initial}
\end{table}

\subsection{Semi-supervised learning}
\subsubsection{Initial Semi-supervised experiment}
\label{Results:Semisupervised_Initial}
Evaluation results of the dataset after evaluating using the pretrained representation are presented in table \ref{bootstrap_1_cycle}, corresponding to the experiment described in Section~\ref{single_class_bootstrap_eval}. Step size refers to the number of the highest confidence predictions per class which are moved from the test set back into the training set. $P_{class}$ refers to the precision of each class $P_{class} = \frac{correct_{class}}{correct_{class} + incorrect_{class}}$ for the images that are to be moved into the training set for that class (which corresponds to the step size). For example, a $P_{non} = 0.999$ with step size = 1000 would indicate that 1 image belonging to the \say{non} class has been misclassified.

\begin{table}[h]
 \begin{center}
 \begin{tabular}{||c|c|c|c|c|c|c|c||} 
 \hline
Method &  Train & Val & Test & Step Size & $P_{cycle}$ & $P_{non}$  & Test Acc \\ [0.5ex] 
 \hline\hline

    \multirow{3}{*}{Transfer} & \multirow{3}{*}{1000} & \multirow{3}{*}{1000} & \multirow{3}{*}{29337} & 100 &  1.0 & 1.0 & \multirow{3}{*}{92\%} \\\cline{5-7}
   &  &  &  & 500 &  1.0 & 1.0 & \\\cline{5-7} 
    &  &  &  & 1000 & 1.0 & 1.0 & \\ 
 \hline
 
    \multirow{3}{*}{Frozen} & 
    \multirow{3}{*}{100} & 
    \multirow{3}{*}{1000} & \multirow{3}{*}{31137} & 100 &  0.8 & 0.94 & \multirow{3}{*}{71\%} \\\cline{5-7}
   &  &  &  & 500 &  0.812 & 0.918 & \\\cline{5-7} 
    &  &  &  & 1000 & 0.796 & 0.909 & \\ 
 \hline

    \multirow{3}{*}{Transfer} & \multirow{3}{*}{5000} & \multirow{3}{*}{1000} & \multirow{3}{*}{21337} & 100 &  1.0 & 1.0 & \multirow{3}{*}{96\%} \\\cline{5-7}
   &  &  &  & 500 &  1.0 & 1.0 & \\\cline{5-7} 
    &  &  &  & 1000 & 1.0 & 1.0 & \\ 
 \hline
 
    \multirow{3}{*}{Frozen} & \multirow{3}{*}{100} & \multirow{3}{*}{100} & \multirow{3}{*}{32937} & 100 &  0.85 & 0.89 & \multirow{3}{*}{69\%} \\\cline{5-7}
   &  &  &  & 500 &  0.824 & 0.89 & \\\cline{5-7} 
    &  &  &  & 1000 & 0.829 & 0.909 & \\ 
 \hline
   \multirow{3}{*}{Frozen} & \multirow{3}{*}{500} & \multirow{3}{*}{500} & \multirow{3}{*}{31337} & 100 & 0.87 & 0.99 & \multirow{3}{*}{75\%} \\\cline{5-7}
   &  &  &  & 500 & 0.894 & 0.984 & \\\cline{5-7} 
    &  &  &  & 1000 & 0.878 & 0.98 & \\ 
 \hline
\end{tabular}
\end{center}
\caption{Results of a single bootstrapping step over different step sizes.}
\label{bootstrap_1_cycle}
\end{table}

Using this data from Table~\ref{bootstrap_1_cycle}, several conclusions can be drawn:

\begin{itemize}
    \item The accuracy of the models built using the Frozen configuration are not suitable for bootstrapping at this level
    \item The size of the validation set only has a minor impact on test set accuracy (2\%) based on the results from the two Frozen experiments with 100 training images. Therefore, 100 validation images may function only slightly worse than 1000, thus further reducing annotation requirements.
\end{itemize}

\subsubsection{Semi-Supervised Consistency}
\label{Results:SemiSupervised_Consistency}

\begin{table}[h]
 \begin{center}
 \begin{tabular}{||c|c|c|c|c|c|c|c||} 
 \hline
Method &  Train & Val & Test & step size & $P_{cycle}$ & $P_{non}$  & Test Acc \\ [0.5ex] 
 \hline\hline
 Transfer & 1000 & 1000 & 29337 & 500 & 1.0 & 1.0 & 92\% \\ 
 \hline
 Transfer & 1500 & 1000 & 28337 & 500 & 1.0 & 1.0 & 93.6\% \\ 
 \hline
 Transfer & 2000 & 1000 & 27337 & 500 & 1.0 & 1.0 & 93.2\% \\ 
 \hline
 Transfer & 2500 & 1000 & 26337 & 500 & 1.0 & 1.0 & 93\% \\ 
 \hline
\end{tabular}
\end{center}
\caption{Results of multiple bootstrapping steps over the single chosen step size}
\label{bootstrap_multiple_cycle}
\end{table}

This section describes the results of the experiment described in Section~\ref{Experiment:SemiSupervised_Consistency}. These results indicate that initializing semi-supervised analysis with around 1000 labeled images per class would result in the ability to consistently improve the accuracy of future iterations of models (based on the $P_{Class}$ results).

However, noticing the trend of test set accuracy introduces another issue: the accuracy increases and then starts decreasing despite more images being present in the training set. Due to the limited size of the testing set, the reduction in the overall size of the test set can be seen to affect evaluation in this case. This is because it becomes increasingly harder to correctly predict from a smaller test set of harder examples, as more confident predictions (i.e., samples that are `easier to predict') are moved out of the test set and harder test-cases are left in. For example, in results in Table~\ref{bootstrap_multiple_cycle}, the test set has shrunk by 3000 images (from 29337 to 26337) corresponding to over 10\% of the test set. These results indicate that a larger test set would be beneficial to further analyse this methodology, additionally allowing exploration into more classes.

\subsubsection{Analysis of multiple classes using Frozen configuration}
\label{results:frozen}

\begin{table}
\begin{center}
 \begin{tabular}{||c|c|c|c|c|c|c||} 
 \hline
Precision & CS & Buildings & GL & Water & Trees & Roads\\ [0.5ex] 
 \hline\hline
 Canberra & 6\% & 100\% & 1\% & 93\% & 90\% * &  100\%\\
 
 \hline
 Ballarat  & 1\% & 100\% & 0\% & 100\% & 100\%  &  100\%\\
 \hline
 Bendigo  & 1\% & 100\% & 0\% & 100\% & 94\% *  &  100\%\\
 \hline
 Cairns & 8\% & 100\% & 2\% & 100\% & 99\% &  100\%\\
 \hline
 Darwin & 0\% & 99\% & 0\% & 100\% &  100\%  &  100\%\\
 \hline
 Geelong & 2\% & 99\% & 3\% & 100\% &  100\%  &  100\%\\
 \hline
 Hobart & 0\% & 99\% & 1\% & 100\% &  100\%  &  100\%\\
 \hline
 Melbourne & 1\% & 100\% & 6\% & 100\% &  100\%  &  100\%\\
 \hline
 Brisbane  & 2\% & 99\% & 5\% & 100\% & 100\%  &  100\%\\
 \hline
 Adelaide  & 10\% & 100\% & 1\% & 100\% & 100\%  &  100\%\\
 \hline
 Toowoomba  & 0\% & 100\% & 1\% & N/A & 100\%  &  100\%\\
 \hline
 Townsville  & 5\% & 100\% & 3\% & 100\% & 100\%  &  100\%\\
 \hline
 Perth  & 1\% & 100\% & 1\% & 100\% & 100\%  &  100\%\\
 \hline
 Wollongong  & 0\% & 100\% & 0\% & 100\% & 100\%  &  100\%\\
 \hline
\end{tabular}
\end{center}
\caption{Results of evaluation across different cities. CS = Cycle Symbols, GL = Green Cycling Lanes. N/A - Detection not present at location,
* - corrupted images detected by model accounted for all erroneous detections.}
\label{multiple_areas_eval}
\end{table}

 This section presents the results of the experiment described under Section~\ref{linearlayer}. The results in Table~\ref{multiple_areas_eval} indicate that more specialized (hence rarer) classes are harder to detect. This is because the potential for misclassification with a dataset of this scale increases when the probability of occurrence of a class decreases. Importantly, this is because more common classes are easier to retrieve as the probability of misclassifying them is lower. As an extreme example, even for a model with perfect accuracy, it would be impossible to retrieve a class which does not exist in the imagery dataset, such as trying to retrieve 'desert' in Antarctica. This experiment concluded that some classes are much harder to detect using self-supervised approaches in datasets of this scale. Therefore, further analysis was focused on improving the performance on such classes, which is discussed under Section~\ref{sec:archival_specialized}. Since results are broadly consistent across different cities (performance is consistent across different columns in Table~\ref{multiple_areas_eval}, indicating that the impact of location is minimal), further analysis was conducted on the entire dataset.
 
\subsubsection{Automated analysis using archival Semi-supervised learning}
\label{results:Archival_semi}

Accuracy results using the semi-supervised method described in Section~\ref{sec:Archival_Initial} are given in Table~\ref{Tab:Semi_acc_results}. As precision (of the category under exploration) is a useful indicator of performance, precision results on the 60-million image dataset can be found under Table~\ref{archival_intial_multitask}. The results are compared against a supervised model trained using the same original training set taken over 5 different runs.

\begin{table}[h!]
 \begin{center}
 \begin{tabular}{||c|c|c|c|c||} 
 \hline
Class & Runs & supervised & semi-supervised & $\Delta$\\ [0.5ex] 
\hline
cycle symbols & 5  & 77.6 $\pm$ 1.32 & 95.3 $\pm$ 1.33 & +17.7\\
basketball courts & 5 & 84.4 $\pm$ 1.39 & 99.8 $\pm$ 0.24 & +15.4\\
solar panels & 5 & 76.0 $\pm$ 1.52 & 99.2 $\pm$ 0.51 & +23.2\\
flat unbuilt & 5 & 99.1 $\pm$ 0.58 & 99.8 $\pm$ 0.24 & +0.7\\
road writing & 5 & 79.7 $\pm$ 4.86 & 98.2 $\pm$ 0.93 & +18.5\\
railway lines & 5 & 74.0 $\pm$ 1.87 & 98.4 $\pm$ 0.66 & +24.4\\
sheep & 5 & 92.3 $\pm$ 4.06 & 99.3 $\pm$ 0.51 & +7.0\\
cycle lanes & 5 & 87.0 $\pm$ 0.71 & 99.6 $\pm$ 0.37 & +12.6\\
road arrows & 5 & 81.4 $\pm$ 3.76 & 96.8 $\pm$ 2.22 & +15.4\\
cars & 5 &  72.7 $\pm$ 2.54 & 95.4 $\pm$ 0.80 & +22.7\\
green cycle lanes& 5 & 67.6 $\pm$ 2.15 & 96.2 $\pm$ 1.12 & +28.6\\
footpaths & 5 & 79.7 $\pm$ 1.36 & 89.7 $\pm$ 1.29 & +10.0\\
buildings & 5 & 81.7 $\pm$ 2.42 & 98.6 $\pm$ 0.58 & +16.9\\
roads & 5 & 82.1 $\pm$ 0.66 & 96.83 $\pm$ 1.70  & +14.7\\
trees & 5 & 77.3 $\pm$ 0.51 & 96.75 $\pm$ 0.75  & +19.4\\
water bodies & 5 & 95.2 $\pm$ 1.6 & 99.625 $\pm$ 0.65 & +4.4\\
pools & 5 & 96.4 $\pm$ 0.8 & 99.5 $\pm$ 0.32 & +3.1\\
sports facilities & 5 & 92.3 $\pm$ 2.29 & 99.9 $\pm$ 0.20 & +7.6\\
\hline
\end{tabular}
\end{center}
\caption{Average validation accuracy (and standard deviation) results over 5 model training runs indicate that the proposed semi-supervised method is more accurate and more consistent}
\label{Tab:Semi_acc_results}
\end{table}

\comment{

0_cycle & 5 & 77.6 $\pm$ 1.32\\
10_basketball & 5 & 84.4 $\pm$ 1.39\\
11_solarpanels & 5 & 76.0 $\pm$ 1.52\\
12_flat_unbuilt & 5 & 99.1 $\pm$ 0.58\\
13_road_writing & 5 & 79.7 $\pm$ 4.86\\
14_railway & 5 & 74.0 $\pm$ 1.87\\
15_sheep & 5 & 92.3 $\pm$ 4.06\\
17_cyclelanes_arch & 5 & 87.0 $\pm$ 0.71\\
18_road_arrows & 5 & 81.4 $\pm$ 3.76\\
1_cars & 5 & 72.7 $\pm$ 2.54\\
2_green_lanes & 5 & 67.6 $\pm$ 2.15\\
3_footpaths & 5 & 79.7 $\pm$ 1.36\\
4_buildings & 5 & 81.7 $\pm$ 2.42\\
5_roads & 5 & 82.1 $\pm$ 0.66\\
6_trees & 5 & 77.3 $\pm$ 0.51\\
7_water & 5 & 95.2 $\pm$ 1.6\\
8_pools & 5 & 96.4 $\pm$ 0.8\\
9_sport & 5 & 92.3 $\pm$ 2.29\\

}

\begin{table}[h!]
 \begin{center}
 \begin{tabular}{||c|c|c||} 
 \hline
Class & Precision(Frozen) & Precision(Archival)\\ [0.5ex] 
 \hline\hline
 Cycle Symbols & 0\% & 85\%\\
 
 \hline
 Green Lanes & 15\% & 98\%\\
 \hline
 Buildings & 100\%  & 100\%\\
 \hline
  Cars & 100\% & 98\%\\
 \hline
 Trees & 100\% & 100\%\\
 \hline
 Water bodies & 100\% & 100\%\\
 \hline
 Solar Panels & 100\% & 100\%\\
 \hline
 Railway Tracks & 62\% & 100\%\\
 \hline
 Footpaths & 94\% & 96\%\\
 \hline
 Lane Arrows on Roads & 44\% & 7\%\\
 \hline
\end{tabular}
\end{center}
\caption{Precision results across multiple tasks on the entire dataset}
\label{archival_intial_multitask}
\end{table}

\subsubsection{Specialized representation learning experiment}
\label{results:specialized_road}
The results in Table~\ref{specialized_results} are compared to the baseline results of this technique from Section~\ref{sec:Archival_Initial} corresponding to the use of a more generic representation. The results indicate that specializing the representation using road imagery has improved performance on categories of infrastructure which coincide with roads, whereas some categories (such as water features) become harder to detect using the proposed methodology.

\aside{
\begin{table}[h]
 \begin{center}
 \begin{tabular}{||c|c|c||} 
 \hline
Class & Precision(Frozen) & Precision(Archival)\\ [0.5ex] 
 \hline\hline
 Cycle Symbols & 0\% & 85\%\\
 
 \hline
 Green Lanes & 15\% & 98\%\\
 \hline
 Buildings & 100\%  & 100\%\\
 \hline
  Cars & 100\% & 98\%\\
 \hline
 Trees & 100\% & 100\%\\
 \hline
 Water bodies & 100\% & 100\%\\
 \hline
 Solar Panels & 100\% & 100\%\\
 \hline
 Railway Tracks & 62\% & 100\%\\
 \hline
 Footpaths & 94\% & 96\%\\
 \hline
 Lane Arrows on Roads & 44\% & 7\%\\
 \hline
\end{tabular}
\end{center}
\caption{Precision results across multiple tasks on the entire dataset}
\label{archival_intial_multitask}
\end{table}
}

\begin{table}[h!]
 \begin{center}
 \begin{tabular}{||c|c|c||} 
 \hline
Class & Precision & Difference from baseline\\ [0.5ex] 
 \hline\hline
 Cycle Symbols & 99\% & 14\% $\uparrow$\\
 \hline
 Green Lanes  & 100\% & 2\% $\uparrow$\\
 \hline
 Buildings  & 99\% & 1\% $\downarrow$\\
 \hline
 Trees & 100\% & 0\%\\
 \hline
 Water bodies & 84\% & 16\% $\downarrow$\\
 \hline
 Solar Panels & 100\% & 0\%\\
 \hline
 Railway Lines & 100\% & 0\%\\
 \hline
 Footpaths & 100\% & 4\% $\uparrow$\\
 \hline
 Lane Arrows on Roads & 99\% & 92\% $\uparrow$\\
 \hline
\end{tabular}
\end{center}
\caption{Results of specializing the frozen representation using road imagery as a first step. It is important to note that urban features correlated with roads have enjoyed an improvement in accuracy, while water bodies(less common near roads) sees a reduction in accuracy.}
\label{specialized_results}
\end{table}

\aside{
\subsubsection{Evaluation of different self-supervision techniques}

\subsubsection{Validation of the use of MoCo for overhead imagery/maps}
Testing on 1667 cities on the city prediction task with no fine-tuning.
V1/V2 refers to the version of MoCo used. 

 \begin{center}
 \begin{tabular}{||c|c|c|c|c|c||} 
 \hline
 Imagery & V1/V2 & Pretrain cities& Pretrain epochs & test cities&  Acc \\ [0.5ex] 
 \hline\hline
 Satellite& V1 & 200 & 200 & 200 & 95\% \\ 
 \hline
 Satellite& V2 & 200 & 200 & 200 & 99\% \\ 
 \hline
 Satellite& V1 & 200 & 200 & 1667 & 81\% \\ 
 \hline
 Satellite& V2 & 200 & 200 & 1667 & 95\% \\ 
 \hline
 
 Satellite& V2 & 1667 & 200 & 1667 & 98\% \\ 
 \hline
 Maps& V1 &1667 & 200 & 1667 & 67\% \\
 \hline
 Maps& V2 &1667 & 200 & 1667 & 64\% \\
 \hline
 Maps& V2 &1667 & 200 & 1667 & 61\% \\
 \hline
\end{tabular}
\end{center}
}

\section{Discussion}

\subsection{Speed and scalability}
With modelling of this scale, it is important to consider how such analysis could be scaled across computing infrastructure to deliver results at speed. The proposed method was able to generate results covering 15 cities in Australia spanning 22,000 km$^2$ and more than 60 million images in 3 hours. This is a throughput of 20 million images per hour or roughly 7,000 square kilometres per hour. These results were generated leveraging the trivial parallelism due to the inherent independent nature of the inference process in neural networks. Processing was performed on 12 V100 GPUs split across 3 nodes (4 GPUs per node) on the Spartan HPC platform\cite{lafayette2016spartan}. On a single GPU, the same workload took 24 hours to complete on a single task. This run-time performance evaluation corresponds to the semi-supervised workflows discussed as part of Section~\ref{results:Archival_semi}.

\subsection{Archival imagery analysis}

\begin{figure}[h!]
\includegraphics[angle=0,scale=0.5]{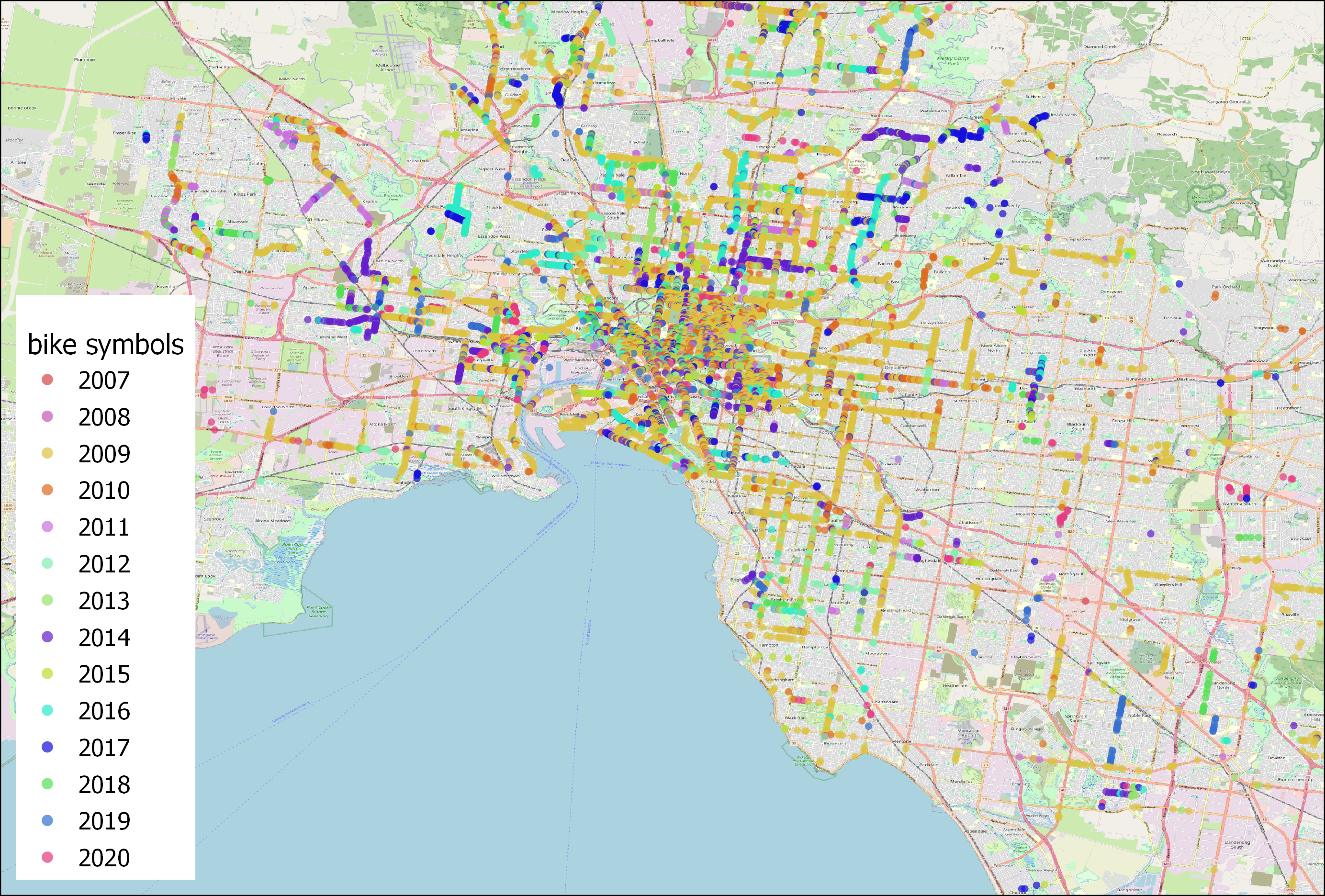}\
\caption{Generated GIS layer of cycling infrastructure over Melbourne}
\label{fig:Melb}
\end{figure}

A straightforward use-case of the models on the task of exploring the evolution of infrastructure over time was used to highlight its utility. Analysis is conducted across the city of Melbourne, and cycling infrastructure over time was analyzed. The first instance of identified infrastructure at a particular location was anotated by the year of detection. This information was used to generate a GIS layer loaded into QGIS which was then visualized as in Fig.~\ref{fig:Melb}. This highlights the utility of the proposed models in providing accurate and consistent data spanning multiple years over a large geographical area. The same data collected by manual processes is laborious to collect and involves repetitive work for the annotator. Beyond providing additional training data, the exploration of archival imagery provides further insights on the growth and change of infrastructure networks.

\subsubsection{Semi-supervised learning mixed with active learning/human-in-the-loop verification}


While the methods introduced in this work attempt fully automated analysis using semi-supervision, there are error rates associated with such analysis. It is possible to lower these error rates in between iterations by performing a manual labelling step on the results to prune out any anomalous detections. These erroneous detections can be quite helpful in directing the model away from such mistakes in future iterations by incorporating these samples into the negative class for the problem at hand. Additionally, even \say{failed} runs where the model cannot provide a high level of accuracy can still be quite useful if the precision is higher than the natural occurrence rate of the detection in the overall dataset. For example, consider a dataset of 100 million images with a detection which occurs in about 0.1\% of images. The dataset would have about 10,000 images containing the class under investigation. If the precision of the generated model is at least 20\% over the top 1000 confidence images, then by annotating those 1000 detections, at least 200 detections will be obtained and can be used to further expand the labeled training dataset. In contrast, it would require manual annotation of at least 200,000 images, on average, to do the same without using a model or some other method of filtering the data.

\subsection{Interpretability}

A key issue in neural network based methods is the interpretability of the generated model. Since the final predictive function the model commits to is the product of multiple complex layers interacting together, it is important to verify that the decision boundary learned by the model is consistent. There are many works in the area of model explainability and interpretability that relate directly to neural networks. Several of these methods were incorporated to provide further validation of our models, by visualizing the activation of the models on input images which contain the corresponding class.

\begin{figure}[h]
\centering
	\subfloat[]{\setlength{\fboxsep}{0pt}\includegraphics[trim = 0mm 0mm 0mm 0mm, clip, height=3.8cm]{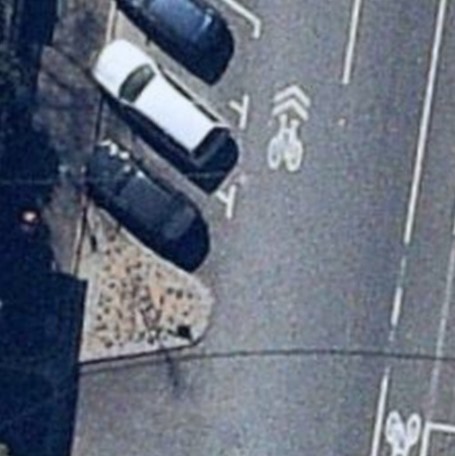}}\,
	\subfloat[]{\setlength{\fboxsep}{0pt}\includegraphics[trim = 0mm 0mm 0mm 0mm, clip, height=3.8cm]{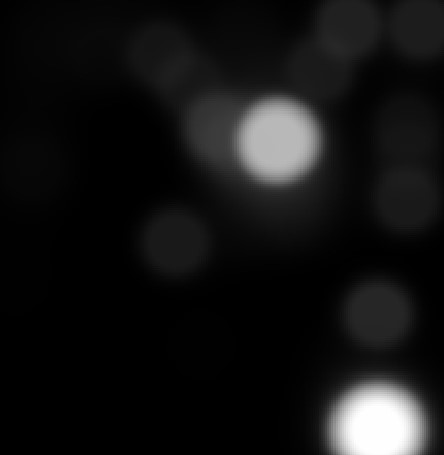}},
    \subfloat[]{\setlength{\fboxsep}{0pt}\includegraphics[trim = 0mm 0mm 0mm 0mm, clip, height=3.8cm]{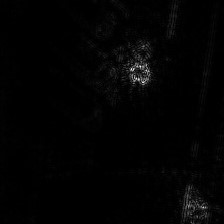}}
	\caption{White activations can be seen in the images (b) (Extremal perturbations) and (c) (Guided backpropagation) corresponding to the activation of the neural network within the image for the original image (a)}
	\label{fig:explainability_activations}
\end{figure}

Two methods (Extremal perturbations \cite{fong2019understanding} and Guided backpropagation \cite{springenberg2015striving}) were used in this regard with the results appearing in Fig.~\ref{fig:explainability_activations}.
\aside{
The following methods were used in this regard:

\begin{itemize}
    \item Extremal perturbations \cite{fong2019understanding}
    \item Guided backpropagation \cite{springenberg2015striving}
\end{itemize}
}

\citet{zhang2018topdown} provide a framework for evaluating attribution techniques by getting the model to "Point" at a single pixel and then scoring based on how far that point is from the given class in the image (15 pixel distance). Points are derived for each technique in a method-dependent fashion.

\begin{figure}[h]
\centering
	\subfloat[]{\setlength{\fboxsep}{0pt}\includegraphics[trim = 0mm 0mm 0mm 0mm, clip, height=3.8cm]{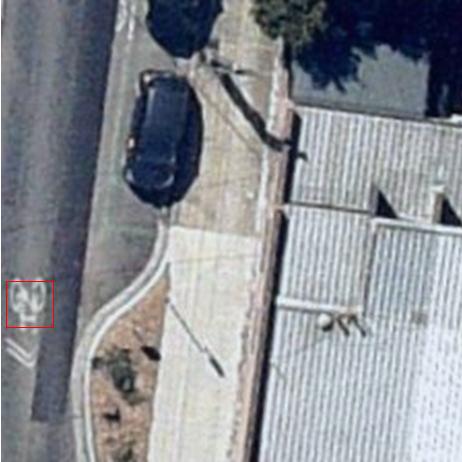}}\,
	\subfloat[]{\setlength{\fboxsep}{0pt}\includegraphics[trim = 0mm 0mm 0mm 0mm, clip, height=3.8cm]{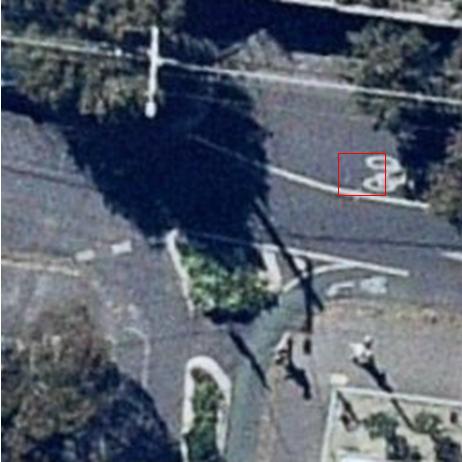}},
    \subfloat[]{\setlength{\fboxsep}{0pt}\includegraphics[trim = 0mm 0mm 0mm 0mm, clip, height=3.8cm]{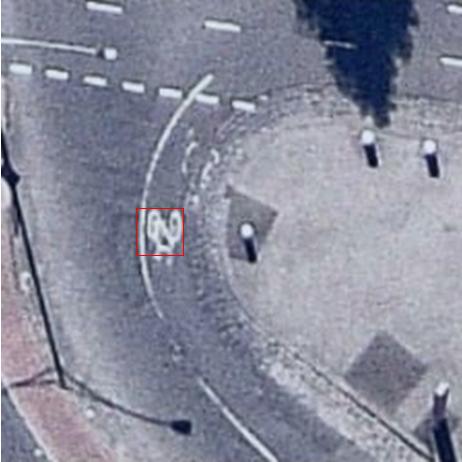}}
	\caption{A red box is drawn around the region within the image most associated with the category under consideration (cycling symbols in this case) by the trained model}
	\label{fig:explainability_eval}
\end{figure}

To generate confidence in the results generated by the neural network models, a similar workflow was implemented using \cite{fong2019understanding}. The single most important image region activated by the neural network was highlighted within the image and manually verified. An example from cycling symbols can be found in Figure \ref{fig:explainability_eval}. Similar results were observed across other classes, however, as this is a class where only a single area within the image corresponds to the task under consideration, this forms one of the harder cases for the model and interpretability technique. Thus, this result was used to highlight and further validate the behaviour of the model.

\subsection{Significance of scalable methods in infrastructure analysis}

Cycling and active transport can address the increasing congestion on road networks from motorised transport, reduce air pollution, and tackle concerning levels of population inactivity. However, cycling is not without risk of injury \cite{henley2012trends} and within increasing numbers of cyclists, comes consequent - though not matched - increases in cycling injury if separated infrastructure is not present \cite{THOMPSON201718}. Specifically, the number of cyclists suffering life-threatening injuries has increased by an average of 7.5\% every year \cite{henley2015trends}. More recently, social distancing measures related to the COVID-19 pandemic have led to an accelerated increase of cycling activity and strong growth in new bicycle sales, globally \cite{fuller2020reactivated}. The promotion and increased uptake of cycling requires investigation into features associated with the accompanying increased numbers of injuries. One of these features is the availability of specific cycling infrastructure, such as marked or physically separated lanes. Our study provides an approach to create such a catalogue of cycling infrastructure, which can have many useful downstream applications such as the development of infrastructure typologies\cite{beck2021typologies}. Importantly, this work showcases how the method can be extended to other types of urban features as well.

\section{Conclusion}

This article proposes a generic method to extract a broad set of features from aerial imagery, which describe the environment in a single image. Although an image segmentation approach can achieve similar results in a single model, one of the major limitations is the requirement of a large amount of samples for model calibration. For example, \citet{azimi2019skyscapes} annotated 31 semantic categories, including low vegetation, tree, paved road, non-paved road, paved parking place, non-paved parking place, bike-way, sidewalk, entrance/exit, and 12 lane-marking types. As user requirements vary, multiple datasets were created by merging some of the detailed categories into higher-level classes (e.g., `nature'). These image segmentation methods have significant potential for  urban infrastructure identification. However, creating annotated training datasets is a highly resource intensive process, with no guarantee that the segmentation categories match the requirements of alternative research questions. In contrast, our method requires only 200 label annotations per category, which is substantially more efficient. Several variations of the methods introduced were also explored, modifying aspects of the self-supervised and semi-supervised learning workflows. Deep learning explainability techniques were applied to verify the hypothesis learned by the model.

This article describes the accuracy of feature detection for various type of infrastructure (e.g., footpaths, cycling lanes), showing commonly encountered infrastructure is easier to detect than rare objects such as cycle symbols. However, the deep learning methods discussed in this article are able to accurately detect any of the investigated types of infrastructure given a sufficient number of training samples. Although the level of initial image annotations can be debated (i.e., set to 200 in this study), a low threshold prevents excessive annotation efforts for features that are easy to distinguish, such as rail tracks. When a higher prediction accuracy is required for certain classes, approaches such as obtaining additional historical imagery at already annotated locations can boost accuracy without the need for further annotation.

\section*{Acknowledgments}

This project was supported by Australian NHMRC Grant GA80134. This research was undertaken using the LIEF HPC-GPGPU Facility hosted at the University of Melbourne. This Facility was established with the assistance of LIEF Grant LE170100200.





\bibliographystyle{elsarticle-num-names}
\bibliography{references.bib}







\end{document}